# A review of predictive uncertainty estimation with machine learning

Hristos Tyralis[1,2,*], Georgia Papacharalampous[3]

[1]Department of Topography, School of Rural, Surveying and Geoinformatics Engineering, National Technical University of Athens, Iroon Polytechniou 5, 157 80 Zografou, Greece (hristos@itia.ntua.gr, https://orcid.org/0000-0002-8932-4997)

[2]Construction Agency, Hellenic Air Force, Mesogion Avenue 227–231, 15 561 Cholargos, Greece (montchrister@gmail.com, https://orcid.org/0000-0002-8932-4997)

[3]Department of Topography, School of Rural, Surveying and Geoinformatics Engineering, National Technical University of Athens, Iroon Polytechniou 5, 157 80 Zografou, Greece (papacharalampous.georgia@gmail.com, https://orcid.org/0000-0001-5446-954X)

*Corresponding author

**Abstract**: Predictions and forecasts of machine learning models should take the form of probability distributions, aiming to increase the quantity of information communicated to end users. Although applications of probabilistic prediction and forecasting with machine learning models in academia and industry are becoming more frequent, related concepts and methods have not been formalized and structured under a holistic view of the entire field. Here, we review the topic of predictive uncertainty estimation with machine learning algorithms, as well as the related metrics (consistent scoring functions and proper scoring rules) for assessing probabilistic predictions. The review covers a time period spanning from the introduction of early statistical (linear regression and time series models, based on Bayesian statistics or quantile regression) to recent machine learning algorithms (including generalized additive models for location, scale and shape, random forests, boosting and deep learning algorithms) that are more flexible by nature. The review of the progress in the field, expedites our understanding on how to develop new algorithms tailored to users' needs, since the latest advancements are based on some fundamental concepts applied to more complex algorithms. We conclude by classifying the material and discussing challenges that are becoming a hot topic of research.

## 1. Introduction

In the vast majority of supervised machine learning applications, point predictions are made that are intended to be close to the actual values of continuous processes. Point predictions are single values that represent the best estimate of a future outcome, based on a set of historical data. They are often used in supervised machine learning applications where the aim is to predict a continuous value, such as the temperature in weather forecasting or the next day's stock price. Machine learning regression algorithms can optimize a squared error (or similar) loss function to issue point predictions. Although point predictions are useful, the information content can be increased when predictions take the form of probability distributions (Dawid 1984, Gneiting and Raftery 2007). In this case, the predictive uncertainty is estimated, and better informed decisions under uncertainty can be made.

Notwithstanding that statistical modelling is mostly associated with inference of parameters, it has been indicated that the ultimate goal should be the prediction of future events (Billheimer 2019). The earliest formal strategy for estimating the probability distribution of predictions consists of fitting a Bayesian statistical model to given data (Roberts 1965, Barbieri 2015). The Bayesian problem is formulated by assigning a prior distribution to the parameters of the statistical model, updating the distribution of the parameters conditional on the data and estimating the predictive uncertainty by integrating the parameters posterior distribution. The probability distribution of the predictions is called predictive distribution, while the procedure can be called predictive uncertainty estimation (or quantification), since probability distributions characterize the uncertainty of the predictions. Although earlier considerations on predictive uncertainty estimation were Bayesian statistical-based for independent and identically distributed (IID) variables, further advancements were made possible due to the relevant progress in time series forecasting, mostly again in Bayesian settings (Chatfield 1996).

In a distinct direction, advancements became also possible due to progress in decision theory through the advent of new loss functions following the theory of consistent scoring functions and proper scoring rules (Gneiting and Raftery 2007, Gneiting 2011a, Dawid and Musio 2014). An indicative example is the quantile loss function (Koenker and Bassett

Jr 1978), which can be used to estimate quantiles of the predictive distribution, when it is minimized in a regression setting.

The machine learning paradigm is grounded on the cross-validation technique, in which the parameters of the model are estimated with respect to minimizing a loss function. There is a consensus that machine learning models are more accurate compared to simple statistical models, when predicting responses due to their flexible nature (Breiman 2001b, Shmueli 2010). Naturally, the advent of new loss functions tailored to estimate probability distributions, combined with the progress in the field of machine learning algorithms, lead to machine learning algorithms that can estimate predictive uncertainty and can be more accurate compared to simpler statistical models.

Here, we review the topic of probabilistic forecasting and prediction using machine learning algorithms. "Prediction" is a general term in the sense that it encompasses the term "forecasting". The latter term refers to the case that one predicts the future conditional on past events, and is regularly accompanied by the concept of temporal dependence. Machine learning algorithms developed to model IID variables can also be applied to forecasting problems in a straightforward way and frequently show good performance, despite not exploiting temporal dependence information. We decided to distinguish the fields of probabilistic prediction and forecasting here, because numerous significant developments have originated in the latter field. Of course, ideas from one field can be transferred directly to the other field; therefore they are introduced in a unifying way hereinafter.

We aim to give an overview of the main concepts, methodologies and research techniques for predictive uncertainty estimation with machine learning algorithms. Moreover, we aim to provide insight on how seemingly unrelated features can be combined to form new algorithms for probabilistic prediction tailored to users' needs, by synthesizing the literature. To this end, we review the following components of probabilistic prediction algorithms:

(a) Concepts related to metrics (consistent scoring functions and proper scoring rules) for assessing probabilistic predictions.

(b) Bayesian settings.

(c) Simpler as well as more complex statistical and machine learning models.

Those components can form new machine learning algorithms, when they are combined.

Furthermore, we examine some special cases of probabilistic prediction including combinations of techniques, time series forecasting, spatial prediction, prediction of extremes and measurement errors. These special cases appear less frequently in the literature (with the exception of combinations of techniques and time series forecasting) and seem to be a subject of timely debate. Challenges in probabilistic prediction are also discussed.

The remainder of the manuscript is structured as follows. Section 2 presents definitions related to probabilistic forecasting and prediction that are associated with Bayesian theory, as well as the definition of the problem of probabilistic prediction in regression settings. Moreover, we list past review papers that surveyed parts of our research topic, and specify how this review paper advances the field and how it differs from the existing literature. The theory of metrics for assessing probabilistic predictions follows in Section 3, while Sections 4 and 5 present statistical and machine learning algorithms for probabilistic prediction, respectively. The metrics section comes before the algorithms section because algorithms are constructed based on loss functions. Section 6 is dedicated to neural networks and deep learning for probabilistic prediction, due to the increasing number of relevant applications using those algorithms. To understand the concepts presented in the aforementioned sections, we demonstrate an example of probabilistic prediction using simulated data in Section 7. Ensemble learning, also termed as combination of predictions in the statistical literature, has been proven to increase predictive performance compared to base learners. We present ensemble learning concepts for probabilistic predictions in Section 8. Special applications of probabilistic predictions that are of interest to the wider scientific community, including among others, temporal, spatial and spatiotemporal prediction, prediction of extreme events, and uncertainty in measurements are presented in Section 9, while related applications in various scientific fields are listed in Section 9.6. The manuscript concludes with a synthesis of results and future outlook in Section 10.

## 2. Definitions and history

### 2.1 Bayesian statistical modelling

Let $\underline{y}$ be the response (dependent) variable of a regression model and let $\underline{x}$ be the $p$-dimensional vector of predictor (explanatory) variables. Note also that $\underline{y}$ and $\underline{x}$ are random variables, while we underlie them to distinguish them from respective observations. A regression model that expresses the relationship between $\underline{x}$ and $\underline{y}$ can be defined by the distribution $F_{\underline{y}|\theta, x}$ of the random variable $\underline{y}$ given a parameter vector $\boldsymbol{\theta}$ and the realization $\boldsymbol{x}$:

$$F_{\underline{y}|\theta, x}(y|\boldsymbol{\theta}, \boldsymbol{x}) := P(\underline{y} \leq y|\boldsymbol{\theta}, \boldsymbol{x}) \tag{1}$$

The prediction problem is then defined as follows: Given the form of the statistical model $F_{\underline{y}|\theta, x}$, realizations $(y, \boldsymbol{x})$ of the vector $(\underline{y}, \underline{x})$ and future yet unobserved realizations $\widetilde{x}$ of $\underline{x}$, estimate the distribution $p(z|y, \boldsymbol{x}, \widetilde{x})$ of future predictions $\underline{z}$ of $\underline{y}$. The distribution $p(z|y, \boldsymbol{x}, \widetilde{x})$ is called predictive distribution (Gelman et al. 2013, Barbieri 2015) and a review of methods for its estimation with machine learning algorithms is the topic of our article.

If $\boldsymbol{\theta}$ was known, then it would be straightforward to estimate the predictive distribution using eq. (1). Unfortunately, in practical situations, that is not the case, consequently the parameter vector $\boldsymbol{\theta}$ has to be estimated from the data. However, a point estimate of $\boldsymbol{\theta}$ (e.g. a maximum likelihood estimate) would ignore the estimation uncertainty, therefore, it seems reasonable to treat $\boldsymbol{\theta}$ as a random variable. Here, Bayesian statistical modelling comes into play. Relevant material regarding predictive inference can be found in some early works in Bayesian statistics (see e.g. the expository study on univariate distributions by Roberts 1965, as well as the summary by Barbieri 2015), while Bayesian theory can be found in a large list of Bayesian books (Robert 2007, Bernardo and Smith 2008, Gelman et al. 2013).

We return to our regression problem by defining the Bayesian statistical regression model made by the statistical model $F_{\underline{y}|\theta, x}(y|\boldsymbol{\theta}, \boldsymbol{x})$ with respective probability density function $f_{\underline{y}|\theta, x}(y|\boldsymbol{\theta}, \boldsymbol{x})$ and a prior distribution on the parameters $p(\boldsymbol{\theta})$. Now given realizations $(y, \boldsymbol{x})$ of the vector $(\underline{y}, \underline{x})$, it is possible to estimate $p(\boldsymbol{\theta}|y, \boldsymbol{x})$, called posterior parameter distribution:

$$p(\boldsymbol{\theta}|y, \boldsymbol{x}) = p(\boldsymbol{\theta}) \, f_{\underline{y}|\theta, x}(y|\boldsymbol{\theta}, \boldsymbol{x}) \, / \int p(\boldsymbol{\theta}) \, f_{\underline{y}|\theta, x}(y|\boldsymbol{\theta}, \boldsymbol{x}) \mathrm{d}\boldsymbol{\theta} \tag{2}$$

Given realizations $(y, \boldsymbol{x})$ and future realizations $\widetilde{x}$ of $\underline{x}$, the predictive distribution of the response variable $\underline{z}$ is:

$$p(z|y, \boldsymbol{x}, \widetilde{x}) = \int f_{\underline{y}|\theta, x}(z|\boldsymbol{\theta}, \widetilde{x}) \, p(\boldsymbol{\theta}|y, \boldsymbol{x}) \, \mathrm{d}\boldsymbol{\theta} \tag{3}$$

There are some notes to be made here:

- Eq. (3) is derived under the assumption of independence of the observations.

- It would be possible to define a prior probability for $\underline{x}$. However, that would not affect the conditional problem (Gelman et al. 2013, Section 14.1).

- Explicit forms for the posterior parameter distribution and the predictive distribution exist in some cases. Still, in the vast majority of cases, those distributions should be simulated with Markov Chain Monte Carlo (MCMC) techniques.

A first consideration on the definition of the problem is related to time series forecasting applications. Indeed, those applications are based on the assumption of temporal dependence of variables; thus, the assumption of independence for deriving Eq. (3) would result in inferior predictive performances. For this specific case, the problem will be reformulated in Section 4.3. Some other questions arising directly from the above definitions are related to the assessment of the predictive performance of different models. The relevant theory will be introduced in Section 3.

## 2.2  Summary of review papers on probabilistic forecasting and prediction

Although the literature on predictions with machine learning is large, the same cannot be said for the topic of probabilistic predictions. A list of books and review papers on subtopics of probabilistic prediction can be found in Table 1. Taking a first look, none work has addressed all subtopics related to probabilistic prediction. Some papers are dedicated to probabilistic prediction with a narrow class of algorithms, such as distributional regression or deep learning. The topic of assessing the predictions is missing from those studies. Vice versa, papers related to scoring probabilistic predictions do not address the topic of algorithms. A large part of the literature examines the topic of forecasting, as well as that of model combinations.

Table 1. Classification of review papers related to probabilistic forecasting and prediction.

| Reference | Title | Metrics – scoring (3) | Bayesian modelling (4.1) | Forecasting (4.3) | Machine learning (4, 5) | Deep learning (6) | Combinations (8) |
|---|---|---|---|---|---|---|---|
| Genest and Zidek (1986) | Combining probability distributions: A critique and an annotated bibliography | | | | | | ✓ |
| Chatfield (1993) | Calculating interval forecasts | | | ✓ | | | |
| Chatfield (1996) | Model uncertainty and forecast accuracy | | | ✓ | | | |
| Winkler (1996) | Scoring rules and the evaluation of probabilities | ✓ | | | | | |
| Hoeting et al. (1999) | Bayesian model averaging: A tutorial | | | | | | ✓ |
| Tay and Wallis (2000) | Density forecasting: A survey | | | ✓ | | | |
| Lampinen and Vehtari (2001) | Bayesian approach for neural networks—review and case studies | | ✓ | | | ✓ | |
| Geweke and Whiteman (2006) | Chapter 1 Bayesian Forecasting | | ✓ | ✓ | | | |
| Gneiting and Raftery (2007) | Strictly proper scoring rules, prediction, and estimation | ✓ | | | | | |
| Robert (2007) | The Bayesian Choice | | ✓ | | | | |
| Casati et al. (2008) | Forecast verification: Current status and future directions | ✓ | | | | | |
| Khosravi et al. (2011) | Comprehensive review of neural network-based prediction intervals and new advances | | | | | ✓ | |
| Wilks (2011) | Chapter 8 - Forecast Verification | ✓ | | | | | |
| Clarke and Clarke (2012) | Prediction in several conventional contexts | | ✓ | ✓ | ✓ | | |
| Vehtari and Ojanen (2012) | A survey of Bayesian predictive methods for model assessment, selection and comparison | | ✓ | | | | |
| Fahrmeir et al. (2013) | Regression: Models, Methods and Applications | | | | ✓ | | |
| Gelman et al. (2013) | Bayesian Data Analysis | | ✓ | | | | |
| Kneib (2013) | Beyond mean regression | | | | ✓ | | |
| Dawid and Musio (2014) | Theory and applications of proper scoring rules | ✓ | | | | | |
| Gneiting and Katzfuss (2014) | Probabilistic forecasting | | | ✓ | | | |
| Carvalho (2016) | An overview of applications of proper scoring rules | ✓ | | | | | |
| Koenker (2017) | Quantile regression: 40 years on | | | | ✓ | | |
| Rue et al. (2017) | Bayesian computing with INLA: A review | | ✓ | | | | |
| Fragoso et al. (2018) | Bayesian model averaging: A systematic review and conceptual classification | | | | | | ✓ |
| Kabir et al. (2018) | Neural network-based uncertainty quantification: A survey of methodologies and applications | | | | | ✓ | |
| Ovadia et al. (2019) | Can you trust your model's uncertainty? Evaluating predictive uncertainty under dataset shift | | | | | ✓ | |
| Bassetti et al. (2020) | Density forecasting | | | ✓ | | | |
| Čížek and Sadıkoğlu (2020) | Robust nonparametric regression: A review | | | | ✓ | | |
| Wood (2020) | Inference and computation with generalized additive models and their extensions | | | | ✓ | | |
| Abdar et al. (2021) | A review of uncertainty quantification in deep learning: Techniques, applications and challenges | | | | | ✓ | |
| Bjerregård et al. (2021) | An introduction to multivariate probabilistic forecast evaluation | | | ✓ | | | |
| Hüllermeier and Waegeman (2021) | Aleatoric and epistemic uncertainty in machine learning: An introduction to concepts and methods | | | | ✓ | | |
| Kaplan (2021) | On the quantification of model uncertainty: A Bayesian perspective | | | | | ✓ | |
| Kneib et al. (2021) | Rage against the mean – A review of distributional regression approaches | | | | ✓ | | |
| Krüger et al. (2021) | Predictive inference based on Markov Chain Monte Carlo output | | ✓ | | | | |
| Makridakis et al. (2021) | The M5 uncertainty competition: Results, findings and conclusions | | | ✓ | | | |
| He et al. (2022) | Risk measures: Robustness, elicitability, and backtesting | ✓ | | | | | |
| Zhou et al. (2022) | A survey on epistemic (model) uncertainty in supervised learning: Recent advances and applications | | | | | ✓ | |
| Gawlikowski et al. (2023) | A survey of uncertainty in deep neural networks | | | | | ✓ | |

Here, we intend to present a complete view of all topics along with their interplay. For instance, loss functions are an essential part of a machine learning algorithm; however, as can be seen from some recent review papers in deep learning (Abdar et al. 2021, Zhou et al. 2022) it is missing from the literature, which focuses on simulation-based techniques, either in simple simulation settings or in Bayesian ones.

Although some recent data competitions have highlighted the importance of machine learning algorithms in efficient forecasting, probabilistic forecasting review papers are relatively old and usually do not include the topic of machine learning. On the other hand, Bayesian textbooks focus on statistical considerations leaving less place to prediction considerations. Model combinations for probabilistic predictions with machine learning and probabilistic predictions focusing on extremes, as well as probabilistic predictions for spatial and spatio-temporal problems, are also missing from the literature.

## 3.   Loss functions for assessing probabilistic predictions

The theory of loss (scoring) functions for point predictions has been extensively surveyed (see e.g. Hyndman and Koehler 2006). Much of the development of machine learning algorithms is due to the knowledge of the properties of loss functions. Therefore, it is natural to ground our study on the theory of scoring functions and scoring rules for probabilistic predictions.

### 3.1   Assessment of quantile predictions

Machine learning practitioners are familiar with the use of loss functions to minimize the error between predicted and observed values for point prediction. Let $L(z, y)$ be a negatively oriented loss (scoring) function that returns a penalty when $z$ is predicted and $y$ realizes. When there are $n$ observations $y_1, ..., y_n$ and $n$ respective point predictions $z_1, ..., z_n$, issued by the regression algorithm then its performance can be scored by averaging the respective penalties:

$$S_n = (1/n) \sum_{i=1}^{n} L(z_i, y_i) \tag{4}$$

As a representative example, let $\rho_{1/2}(u)$ be half the absolute function

$$\rho_{1/2}(u) = |u|/2 \tag{5}$$

and $L_{1/2}(z, y)$

$$L_{1/2}(z, y) = \rho_{1/2}(y - z) \tag{6}$$

be half the absolute error (AE) function. A machine learning algorithm trained by minimizing $L_{1/2}(z, y)$, predicts the median of the probability distribution of the response variable (Gneiting and Raftery 2007).

Similarly to predicting the median, one may receive a directive to predict a quantile of the probability distribution of the response variable. Let, $Q_{\underline{y}}(\alpha)$ defined by

$$Q_{\underline{y}}(\alpha) := F_{\underline{y}}^{-1}(\alpha), \, 0 \leq \alpha \leq 1 \tag{7}$$

be the $\alpha^{\text{th}}$ quantile of $\underline{y}$. The quantile is a functional of the probability distribution. Intuitively, $100 \, \alpha\%$ of the values are lower than $Q_{\underline{y}}(\alpha)$, while one could predict quantiles at a dense grid of quantile levels, to estimate the predictive distribution. A special case of eq. (7) is the median of $\underline{y}$, $Q_{\underline{y}}(1/2)$.

Expanding the definition of the quantile to the regression setting, let $Q_{\underline{y}|\boldsymbol{x}}(\alpha|\boldsymbol{x})$

$$Q_{\underline{y}|\boldsymbol{x}}(\alpha|\boldsymbol{x}) := F_{\underline{y}|\boldsymbol{x}}^{-1}(\alpha) \tag{8}$$

be the $\alpha^{\text{th}}$ quantile of $\underline{y}$ conditional on $\boldsymbol{x}$. Now, assume that one predicts the value $z$ for $Q_{\underline{y}}(\alpha)$. Let also $y$ be the respective realization of $\underline{y}$. In the familiar case of predicting the median of $\underline{y}$, the prediction's score would be $L_{1/2}(z, y)$ with optimal score being 0, when $z = y$. The absolute error function is generalized by the tilted absolute value function (Koenker and Bassett Jr 1978, Gneiting 2011a), defined by

$$\rho_\alpha(u) := u \, (\alpha - \mathbb{I}(u \leq 0)) \tag{9}$$

Here $\mathbb{I}(\cdot)$ denotes the indicator function and $\alpha$ is the quantile level of interest. For $\alpha = 1/2$, eq. (9) reduces to eq. (5). The tilted absolute value function is positive and negatively oriented, i.e. the objective is to minimize it, and equals to 0, when $u = 0$. Let

$$L_\alpha(z, y) := \rho_\alpha(y - z) \tag{10}$$

be the quantile loss function (or asymmetric piecewise linear scoring function) that returns the penalty $L_\alpha(z, y)$ to the prediction $z$ of the $\alpha^{\text{th}}$ quantile of $\underline{y}$ when $y$ realizes (Gneiting 2011a).

While, intuitively, a point prediction $z$ for the median of $\underline{y}$ would be considered satisfying if the penalty $L_{1/2}(z, y)$ is low, the extension to the quantile loss function is not straightforward. For understanding the properties of the quantile loss function, we illustrate in Figure 1, $L_\alpha(z, 0)$ for varying predictive quantiles $z$ at the quantile levels $\alpha \in \{0.05, 0.95\}$. The expectile loss function illustrated in the same figure will be explained later in Section 3.2. Assuming that $z > 0$, it is clear that $L_{0.95}(z, 0) < L_{0.95}(-z, 0)$ and $L_{0.05}(z, 0) > L_{0.05}(-z, 0)$, while the quantile loss function is asymmetric. The asymmetry is due to assigning different weights to the error $y - z$ depending on its sign and allows estimating quantiles at levels different from ½, assuring that the correct ratio of observations $100\alpha\%$ lies below the prediction (Koenker and Hallock 2001). At level ½, $L_{1/2}(z, y)$ becomes symmetric; see eq. (5).

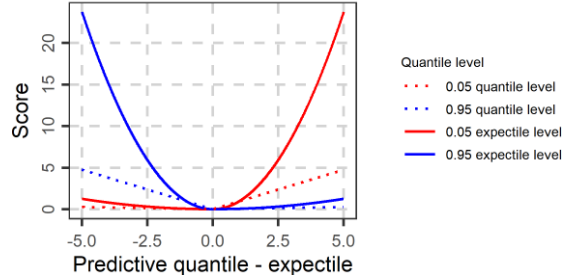

Figure 1. Illustration of the quantile and expectile scores at the quantile levels $\alpha \in \{0.05, 0.95\}$ and expectile levels $\tau \in \{0.05, 0.95\}$, respectively, when $y = 0$ realizes and for varying predictive quantiles and expectiles $z$ (see eqs. (10) and (14)).

A favourable property of the quantile loss function is that it is strictly consistent for the quantile of probability distributions with finite first moments (Murphy and Daan 1985, Gneiting 2011a). That means that following a directive to predict a functional of the probability distribution (e.g. a quantile), a consistent scoring function will optimise the modeller's expected score (Ehm et al. 2016, Taggart 2022a).

In particular, a scoring function $L: D \times D \to [0, \infty)$, where D is the potential range of outcomes of variable $\underline{y}$, is consistent for a functional $T$ that maps $\mathcal{F} \to T(\mathcal{F}) \subseteq D$, relative to the class $\mathcal{F}$ of probability distributions if

$$\mathrm{E}_{\mathcal{F}} L(t, \underline{y}) \leq \mathrm{E}_{\mathcal{F}} L(z, \underline{y}) \tag{11}$$

for all probability distributions $F \in \mathcal{F}$, all $t \in T(\mathcal{F})$ and all $z \in D$. $L$ is strictly consistent if it is consistent and equality of the expectations implies that $z \in T(\mathcal{F})$ (Gneiting 2011a). A functional is elicitable if there exists a scoring function that is strictly consistent for it (Gneiting 2011a).

It is proved that $L$ is consistent for the $\alpha^{\text{th}}$ quantile, if and only if is of the form

$$L_\alpha(z, y) = (g(y) - g(z)) \ (\alpha - \mathbb{I}(y - z \leq 0)) \tag{12}$$

where $g$ is a nondecreasing function on $D$. If $g$ is strictly increasing, then $L$ is strictly consistent. $L$ is called generalized piecewise linear (GPL) of order $\alpha$ (Raiffa and Schlaifer 1961, p.196, Gneiting 2011a), while $L_\alpha(z, y)$ is a special case of the GPL. Under the GPL, any $\alpha^{\text{th}}$ quantile is an optimal point prediction (Gneiting 2011b). An intuitive explanation of the GPL scoring function is that $g$ expresses some utility and that the loss by predicting $z$ when $y$ realizes is asymmetrically (depending on the sign of the difference) proportional to $|g(y) - g(z)|$ (Gneiting [2011b]). Some further results on the interpretation of scoring functions that are consistent for quantiles as well as the use of plots for the comparisons of competing predictions can be found in Ehm et al. ([2016]).

## 3.2 Assessment of expectile predictions

Expectiles, introduced by Newey and Powell ([1987]), are functionals of the probability distribution. They are least squares analogues of quantiles, i.e. they are generalizations of the mean. Similar to eq. (7), the expectile $e_\tau$ at the expectile level $\tau$ can be defined by inverting the following equation:

$$\tau = \mathrm{E}[|\underline{y} - e_\tau|\; \mathbb{I}(\underline{y} \le e_\tau)] \,/\, \mathrm{E}[|\underline{y} - e_\tau|)] \tag{13}$$

From eq. (13) it is clear that $e_\tau$ is such that the mean distance from all $\underline{y}$ below $e_\tau$ is 100 $\tau\%$ of the mean distance between $\underline{y}$ and $e_\tau$ (Daouia et al. [2018]) and that $e_{1/2}$ is the mean of the probability distribution of $\underline{y}$.

The following asymmetric piecewise scoring function is strictly consistent for the expectile $e_\tau$ and has similar behaviour with the quantile loss function; see Figure 1 (Gneiting [2011a]).

$$L_\tau(z, y) := (y - z)^2 \,|\tau - \mathbb{I}(y - z \le 0)| \tag{14}$$

When $\tau = \frac{1}{2}$, eq. (14), reduces to half the squared error function

$$L_{1/2}(z, y) := ((y - z)^2)/2 \tag{15}$$

It is proved that a scoring function is consistent for the $\tau^{\text{th}}$ expectile, if and only if is of the form

$$L_\tau(z, y) = (h(y) - h(z) - h'(z)\,(y - z))\,|\tau - \mathbb{I}(y - z \le 0)| \tag{16}$$

where $h$ is a convex function with subgradient $h'(z)$ (Gneiting [2011a]). If $h$ is strictly convex, then it is strictly consistent for the $\tau^{\text{th}}$ expectile. For some further results on the interpretation of scoring functions that are consistent for expectiles as well as the use of plots for the comparisons of competing predictions, as well as an interpretation of expectiles as a risk measure, see Ehm et al. ([2016]).

Expectiles are useful risk measures in econometrics, since they are both elicitable and coherent (quantiles are not coherent risk measures) (Bellini et al. [2014], Emmer et al. [2015]). Furthermore, they can be informative regarding the size of losses, while quantiles are informative regarding the frequency of losses (Taylor [2021]). Despite their favourable properties, expectiles are used less frequently compared to their counterparts (quantiles) possibly due to issues related with their interpretability (Waltrup et al. 2015). A relationship exists between quantiles and expectiles (Jones 1994). For very heavy-tailed distributions (with tail index larger than ½), extreme expectiles are larger than extreme quantiles in magnitude, while for probability distributions with tail index lower than ½, the opposite is true (Bellini et al. 2014).

For a generalization of quantiles and expectiles, see *M*-quantiles in Breckling and Chambers (1988), Koltchinskii (1997), with related consistent scoring functions (Huber quantile scoring functions) defined by Taggart (2022b) while also some intermediate loss functions between the quantile and the expectile losses exist, called $k^{th}$ power expectile losses (Jiang et al. 2021).

## 3.3   Proper scoring rules

Assume that a modeller predicts the full distribution $P \in \mathcal{P}$ for the response variable $\underline{y}$, while $F$ is the true predictive distribution. Let $y$ be a future realization of $\underline{y}$. Similarly to the point prediction case, the predictive distribution could be assessed based on the pair $(P, y)$ (Dawid 1984). To assess the prediction, we define the scoring rule $S(P, y)$ (with negative orientation) that returns a score dependent on $P$ and $y$, while the expected score under $F$ is (Gneiting and Raftery 2007):

$$s(P, F) := \int S(P, y) \mathrm{d}F(y) \tag{17}$$

Note that the scoring rule resembles a scoring function, as an error measure, with difference being that its first argument is a distribution instead of a point. The scoring rule $S$ is proper relative to the class of distributions $\mathcal{P}$ if

$$s(F, F) \leq s(P, F) \; \forall \; P, F \in \mathcal{P} \tag{18}$$

and strictly proper when $s(F, F) = s(P, F)$ if and only if $P = F$ (Winkler and Murphy 1968, Gneiting and Raftery 2007), i.e. a scoring rule is proper if it optimizes the expected score, when $y$ from $F$ realizes and the modeller issues a probabilistic prediction $F$ (Gneiting and Raftery 2007). The divergence $d(P, F)$ is defined by:

$$d(P, F) := s(P, F) - s(F, F) \; \forall \; P, F \in \mathcal{P} \tag{19}$$

Interpreting $d(P, F)$ as a divergence is meaningful for strictly proper scoring rules (Bröcker 2009). In addition, the entropy $e(P)$ is defined by

$$e(P) := s(P, P) \; \forall \; P \in \mathcal{P} \tag{20}$$

The entropy $e(P)$ can be interpreted as lack of information (Bröcker 2009).

Minimizing the expected score, will lead to the true predictive distribution, regardless of the type of the scoring rule, although finite samples may affect the efficiency of the scoring rules (Loaiza-Maya et al. 2021).

### 3.3.1 Decomposition of proper scoring rules

A proper scoring rule can be decomposed into different components that contribute to the overall score of a probabilistic forecast, and these components can help us understand the properties of the probabilistic forecast. A general decomposition has been proposed by Bröcker (2009). Assuming that $\bar{\pi}$ is the probability distribution function of $\underline{y}$ (also termed climatology) and $\pi_{(\gamma)}$ is the conditional probability of $\underline{y}$ given the forecasting scheme $\gamma$, $\mathrm{E}_{\mathcal{F}}S(\gamma, \underline{y})$ can be decomposed into three components:

$$\mathrm{E}_{\mathcal{F}}S(\gamma, \underline{y}) = e(\bar{\pi}) - \mathrm{E}_{\mathcal{F}}d(\bar{\pi}, \pi_{(\gamma)}) + \mathrm{E}_{\mathcal{F}}d(\gamma, \pi_{(\gamma)}) \tag{21}$$

Here $e(\bar{\pi})$ is called the uncertainty of $\underline{y}$, $\mathrm{E}_{\mathcal{F}}d(\bar{\pi}, \pi_{(\gamma)})$ is called resolution (or sharpness) and $\mathrm{E}_{\mathcal{F}}d(\gamma, \pi_{(\gamma)})$ is called reliability.

The uncertainty is the expectation of the score of a climatology forecasting scheme. In the literature of point forecasting, the climatology is a forecasting scheme that is equal to the mean of dependent variable and can be used as the simplest benchmark. The resolution is a form of variance of $\pi_{(\gamma)}$. Larger resolution means a better score. We have a reliable forecast when $\gamma = \pi_{(\gamma)}$, therefore the reliability is the deviation of $\gamma$ from $\pi_{(\gamma)}$.

### 3.3.2 Continuous Ranked Probability Score (CRPS)

Probabilistic predictions should be sharp and well calibrated. Sharpness refers to the concentration of the distribution, with higher concentration being preferable. Calibration refers to the statistical consistency between the probabilistic predictions and the observations, in the sense that events that are predicted to occur with probability $p$, should be realized with frequency $p$. Sharpness is a property of the predictions while calibration is a property of both predictions and observations. Sharpness of the predictive distribution should be maximized subject to calibration, while a variety of proper scoring rules has been determined towards this direction (Gneiting et al. 2007).

Amongst the variety of proper scoring rules, the most widely used is the Continuous Ranked Probability Score (CRPS), defined by (Epstein 1969, Matheson and Winkler 1976,

Gneiting and Raftery 2007)

$$\mathrm{CRPS}(F, y) := \int_{-\infty}^{\infty} (F(z) - \mathbb{I}(y - z \leq 0))^2 \, \mathrm{d}z \tag{22}$$

Advantages of the CRPS is that is defined directly in terms of cumulative distribution functions thus it allows assessing forecasts in terms of samples (e.g. simulations) (Gneiting and Raftery 2007). Assuming that a simulation of $K$ ensembles from a Bayesian statistical model, represents the predictive distribution, then the CRPS of the prediction can be written as a sum of quantile scores (see eq. (10)) applied to each of the $K$ ensemble members, where the level of $k^{\mathrm{th}}$ lower ensemble member should be set equal to $(k - 0.5)/K$ (Bröcker 2012). Explicit expressions of the CRPS for given $P$ and $y$ or for ensemble (simulated) predictions as well as a related software implementation can be found in Jordan et al. (2019) and Krüger et al. (2021). For ordered ensemble members $z_{(i)}$ with $z_{(1)}$ < ... < $z_{(K)}$ the CRPS can be estimated by

$$\mathrm{CRPS}(z_1, ..., z_k; y) = (2/K^2) \sum_{i=1}^{K} (z_{(i)} - y) \, (K \, \mathbb{I}(y - z_i < 0) - i + 1/2) \tag{23}$$

while the relevant expression for unordered ensembles in not recommended due to computational reasons.

Some further considerations from the theory of proper scoring rules follow. The CRPS can be decomposed into reliability, resolution and uncertainty (Hersbach 2000); see also Section 3.3.1, while probabilistic predictions for circular quantities can be can be assessed using a CRPS variant (Grimit et al. 2006). The CRPS generalizes the absolute error when the predicted distribution is a point measure, while an approximate relationship exists between the CRPS and the Root Mean Squared Error (RMSE, Leung et al. 2021).

### 3.3.3 Scoring rules for quantiles

As noted in Section 3.1, a probabilistic prediction can take the form of multiple predictive quantiles $z_1, ..., z_k$ at levels $\alpha_1, ..., \alpha_k$. Then a proper scoring rule for assessing the predicted quantiles at the given levels is of the form

$$S(z_1, ..., z_k; y) = - \sum_{i=1}^{k} [\alpha_i \, s_i(z_i) \, + \, (s_i(y) - s_i(z_i)) \, \mathbb{I}(y - z_i \leq 0)] + h(y) \tag{24}$$

where $s_i(\cdot)$, $i = 1, ..., k$ is non-decreasing and $h$ is an arbitrary function (Gneiting and Raftery 2007). Setting $k = 1$, $s_1(y) = y$ and $h(y) = \alpha \, y$ the quantile proper scoring rule reduces to the quantile scoring function. The quantile score can be decomposed in reliability, resolution and uncertainty (Bentzien and Friederichs 2014).

### 3.3.4 Scoring rules for intervals

Given quantile levels $\alpha_1$ and $\alpha_2$, with $\alpha_1 < \alpha_2$, a $\alpha_2 - \alpha_1$ a prediction interval $[q_1, q_2]$, corresponds to quantiles $q_1$ and $q_2$ of the response variable at levels $\alpha_1$ and $\alpha_2$, respectively. A central prediction interval $1 - \alpha$, corresponds to $\alpha_1 = \alpha/2$ and $\alpha_2 = 1 - \alpha/2$. By setting $\alpha_1 = \alpha/2$, $\alpha_2 = 1 - \alpha/2$, $s_1(y) = 2\,y/\alpha$, $s_2(y) = 2\,y/\alpha$, and $h(y) = 2\,y/a$ in eq. (24), the central interval scoring rule is defined by (Dunsmore 1968, Winkler 1972, Gneiting and Raftery 2007)

$$S(z_1, z_2; y) = (z_2 - z_1) + (2/\alpha)\,(z_1 - y)\,\mathbb{I}(y - z_1 < 0) + (2/\alpha)\,(y - z_2)\,\mathbb{I}(z_2 - y < 0) \quad (25)$$

We note that we focus on prediction intervals for a single observation (Chatfield 1993). A possible requirement for a prediction interval is that it should have a specified coverage probability, i.e. a specified frequency of future observations falling within the prediction interval (Christoffersen 1998). Research has been done towards this direction (Chudý et al. 2020). Beyond considerations, on how this coverage probability should be set (Landon and Singpurwalla 2008), Askanazi et al. (2018) state that the problem of comparing prediction intervals when the coverage probability is specified but the quantiles are not, is difficult and perhaps unsolvable.

More generally, it has been shown that the shortest prediction interval and those intervals determined by an endpoint or midpoint are not elicitable, unless an endpoint is given via a quantile (Brehmer and Gneiting 2021, Fissler et al. 2021). Furthermore, the equal-tailed interval and the modal interval are elicitable (Brehmer and Gneiting 2021), while the mode and modal intervals are not indirectly elicitable (Dearborn and Frongillo 2020). Finally, it is possible to construct and evaluate expectile-bounded prediction intervals, similarly to quantile-bounded ones (Taylor 2021).

## 3.4 Some more proper scoring rules for assessing probabilistic predictions

A comprehensive list of proper scoring rules can be found in Gneiting and Raftery (2007) and Dawid and Musio (2014), including the log score (Good 1952), the Tsallis score (Tsallis 1988), the Brier score (Brier 1950), the Bregman score, the survival score, the Hyvärinen score (Hyvärinen and Dayan 2005), the composite score, the pseudo score, the diagonal score (Bouallègue et al. 2018), the energy score (Gneiting and Raftery 2007) and the variogram-based proper scoring rules (Scheuerer and Hamill 2015). Of special interest is the log score defined by

$$S(P, y) = -\log(p(y)) \quad (26)$$

The log score is the only proper scoring rule that is local, i.e. it depends on the predictive distribution only through its value at the event $y$ that realizes (Gneiting and Raftery 2007, Dawid and Musio 2014), thereby ignoring probabilities of events that could have happened but did not (Krüger et al. 2021); see also proper local scoring rules of order $k$ in Ehm and Gneiting (2012) and Parry et al. (2012). The log score is connected with likelihood inference in statistical modelling. In comparison, for instance, with the CRPS, a major drawback of the log score is that is restricted to predictive densities (Gneiting and Raftery 2007). Still, there are also advantages of local proper scoring rules. For such advantages, see the discussion in Du (2021).

Some other proper scoring rules are the error-spread scoring rule, which is formulated with respect to moments of the predictive distribution (Christensen et al. 2015, Christensen 2015), a scoring rule for simultaneous events (Grant et al. 2019), a proper scoring rule motivated by the form of Anderson–Darling distance of distribution functions (Barczy 2022), threshold-and quantile-weighted scoring rules (Gneiting and Ranjan 2011), scoring rules that exploit temporal dependence between events (Lai et al. 2011), joint scoring rules (Lichtendahl Jr. and Winkler 2007) and the asymmetric continuous probabilistic score (ACPS) (Iacopini et al. 2022).

## 3.5 Some more related concepts

Most score values are scale depended (i.e. they depend on the magnitude of observations), therefore skill scores may be used instead to compare predictions for multiple cases (e.g. for multiple time series). Skill scores are standardized version of scoring rules, but they are not necessarily proper (Gneiting and Raftery 2007).

Arbitrary scoring rules can be made proper, while details on such constructions (called properizations) can be found in Dawid (2007) and Brehmer and Gneiting (2020). Tailored scoring rules to client's specified requirements should be selected among the list of scoring rules presented in the previous sections, while it is possible to construct new scoring rules according to requirements of prediction users; see e.g. Johnstone et al. (2011), Machete (2013) and Merkle and Steyvers (2013).

Finally, Liu et al. (2020a) argue that minimizing a proper scoring rule over the input space is not possible, because the true distribution of data cannot be learned by a model, therefore they proposed minimizing a scoring rule with respect to all possible true distributions, transforming the minimization to a minimax problem.

In economic and financial sciences, backtesting is used to assess the performance of a model when predicting a time series. For example, when predicting Value at Risk (VaR), which is a quantile of the predictive distribution, one is interested in ensuring that the conditional coverage of predictions is equal to the nominal level of VaR. Misspecification of the conditional coverage is termed as a violation. A review of backtesting methods for VaR can be found in Zhang and Nadarajah (2018) who have grouped the backtesting methods into four categories: unconditional test methods (e.g. Probability of Exceedance (POF) test by Kupiec 1995), conditional test methods, independence property test methods (e.g., the dynamic quantile (DQ) test by Engle and Manganelli 2004) and other test approaches. Unconditional methods focus on the unconditional coverages, while independence property test methods evaluate the extent to which a VaR measure's performance is independent from one period to the next (Zhang and Nadarajah 2018). The concept of elicitability presented earlier regards comparison of multiple models, while backtesting refers to the validation of a single model (He et al. 2022).

## 4.    Early history and simple models

Before proceeding to machine learning models, a small overview of simpler models that are appropriate for probabilistic predictions may serve in understanding how more complex and accurate models can be built. In particular, we will present the theory of the simple Bayesian statistical ordinary linear regression model, the linear in parameters quantile regression, as well as an overview of time series and copula models. Much of the theory of machine learning models is based on simpler models, with added complexity resulting in improved performance.

### 4.1   Bayesian statistical models

Returning to the Bayesian statistical models and following eq. (3), it is proved that the predictive distribution for the Gaussian ordinary linear regression model with unknown parameters, when assigned a uniform noninformative prior distribution to ($\beta$, $\log\sigma$), where $\sigma$ is the variance of the error term $\varepsilon_i$

$$y_i = \beta_1 x_{i1} + ... + \beta_p x_{ip} + \varepsilon_i, \, i = 1, ..., n \tag{27}$$

is multivariate Student (Gelman et al. 2013, Chapter 14) with $n - p$ degrees of freedom, location parameter $\widetilde{X} u$ and squared scale matrix $s^2 (I + \widetilde{X} V_u \widetilde{X}^{\mathrm{T}})$, where $\widetilde{X}$ is the matrix of future $\widetilde{x}$ values, $X$ is the $n \times p$ matrix of predictors (recall from Section 2.1 that $n$ is the number of samples), $y$ is the $n \times 1$ vector of responses and

$$u = (X^T X)^{-1} X^T y \tag{28}$$

$$V_u = (X^T X) \tag{29}$$

$$s^2 = (1/(n-k)) (y - X u)^T (y - X u) \tag{30}$$

Obviously, it is possible to estimate the predictive distribution for the linear regression, with MCMC simulations (Gelman et al. 2013, Chapter 14). Early considerations of predictive inference for Gaussian models include Geisser (1965), Chew (1966) and Eaton et al. (1996), among others. Predictive distributions for Bayesian statistical models beyond the Gaussian case may not be defined explicitly; therefore, simulations in a MCMC setting may be appropriate and perhaps the only means to proceed.

Although Bayesian statistical modelling is mostly concerned with in-sample performances, where it is essential to specify correctly the data generating process, cross-validation (with out-of-sample performance) using various loss functions has also been the subject of recent research. For relevant examples, see Vehtari and Ojanen (2012), Piironen and Vehtari (2017) and Vehtari et al. (2017). Problems related to cross-validation schemes are that Bayesian statistical models are mostly applied to small datasets, while as the dataset increases, the computational cost of cross-validation may also increase significantly.

To overcome computational costs related to the increasing size of datasets and intractability of likelihoods, Approximate Bayesian Computation (ABC) has been proposed as a solution. ABC is an algorithm that simulates artificial data using parameters from the prior distribution and keeps the data that are close to the observed data according to a distance metric. ABC methods for estimating predictive distributions has been surveyed by Frazier et al. (2019) and has been found to provide efficient predictions that are nearly identical, although inferior, to exact ones.

Focused Bayesian prediction (Loaiza-Maya et al. 2021) is a method for Bayesian predictions, in which correct specification of the data generation process is not required. Updating is done using a specified measure (e.g. a proper scoring rule), while the method is also applicable to improve combinations of algorithms (see also Section 8 for an exposition of combinations of algorithms).

## 4.2 Quantile and expectile regression linear in parameters

### *4.2.1 Quantile regression*

As mentioned in Section 3.1, quantile predictions can be assessed using the quantile loss function defined by eq. (10). Therefore, the idea to estimate the parameters of the linear regression model defined by eq. (27), elaborated by Koenker and Bassett Jr (1978), is natural when one's model is intended to predict quantiles. Optimization of the model is done by averaging the losses over the sample, similarly to what is made for least squares regression. Quantile regression is appropriate when (Waldmann 2018): (a) one is interested in predicting functionals beyond the mean; (b) the parametric form of the predictive distribution is now known; (c) outliers exist (in which case quantile regression is robust); and (d) in presence of heteroscedasticity (i.e. when the variance depends on the covariates). Several developments of quantile regression have been proposed. Although we cannot be exhaustive on them, the following may be of interest, while a brief exposition can be found in Koenker (2017) and an extensive treatment can be found in Koenker et al. (2017).

**Regularization in quantile regression**: In cases where the estimation of the model's parameters is unstable, a regularization process can be applied to improve inference (Bickel and Li 2006). This can be made, for instance, by using ridge regression (Hoerl and Kennard 1970) or lasso (Tibshirani 1996), among other methods for achieving prediction improvements, which may include models with many parameters or high dimensional problems. Regularization has been implemented in quantile regression models, as it is described in overviews such as those by Mizera (2017), Wang (2017) and Belloni et al. (2017). A characteristic example of such quantile regression models is the lasso estimator by Ye and Padilla (2021).

**Bayesian statistical modelling and quantile regression**: Bayesian statistical models for quantile regression can be useful in cases where inference on the parameters is required in the usual Bayesian settings, or in cases where the optimization procedure is difficult. The specification of parametric likelihoods is not possible in quantile regression methods; thus, working likelihoods are usually implemented. An overview of methods on Bayesian quantile regression can be found in Wang and Yang (2017), while the first relevant study appears to be Yu and Moyeed (2001).

**Multivariate quantiles**: Quantile regression can be extended to the multiple-output case. Assuming that quantiles of $d \geq 2$ variables have to be predicted simultaneously, the extensions are natural, although the problem's complexity increases considerably. An overview of methods can be found in Hallin and Šiman (2017). Approaches to the problem include the directional one, in which the multivariate problem is reduced to several univariate ones, in which the focus is on the marginal distributions (see e.g. Hallin et al. 2010). Another approach is the direct one, with the following alternatives: (α) Extension of the loss function to cover the spatial case (spatial loss function); (b) the concept of elliptical quantiles; and (c) the concept of depth-based quantiles. An example for estimating simultaneously multiple quantiles is proposed by Firpo et al. (2021) using a generalized method of moments (GMM).

**Quantile crossing problem**: In the usual case, conditional quantiles in quantile regression are estimated independently, i.e. by fitting different models at varying quantile levels. Consequently, it is possible that a predicted quantile at a given level may be higher than a predicted quantile at a higher level. That problem of quantile regression algorithms is called quantile crossing. Several methods have been proposed to remedy the quantile crossing problem, e.g. by sorting the estimated quantiles (Chernozhukov et al. 2010) or by enforcing non-crossing constraints (Takeuchi et al. 2006).

**Survival analysis with quantile regression**: In survival analysis, the duration until one event occurs is of interest. Censoring in survival analysis refers to the case that the time to event cannot be observed, for example, because the study may stop before the event realizes. Overviews of methods for survival analysis including the censoring case can be found in Ying and Sit (2017), Li and Peng (2017) and Peng (2017, 2021), while the first relevant study appears to be Powell (1986).

**Semiparametric models**: Quantile regression for semiparametric models (i.e. regression models that combine parametric and nonparametric models) has been proposed by Waldmann et al. (2013). Components of the models may include nonlinear effects, spatial effects, non-normal random effects and more. Due to optimization difficulties, such models are formulated in Bayesian settings.

**Panel analysis**: Quantile regression has been applied to panel data. For instance, Geraci and Bottai (2014) modelled multiple random effects.

**Optimization procedures**: An important problem of quantile regression algorithms is that the quantile loss function is not everywhere differentiable; therefore, gradient based optimization methods are not always applicable. For efficient achieving optimizations, approximations to the quantile loss function have been employed. A relevant example can be found in Zheng (2011).

### 4.2.2 Expectile regression

Similarly to what applies to quantile regression, the parameters of the linear regression model, defined by eq. (27), can be estimated by minimizing the expectile loss function defined by eq. (14). Expectile regression has been elaborated by Newey and Powell (1987) and has similar properties with quantile regression when contrasted to Bayesian statistical models. On the other hand, there are some differences between quantile and expectile regression, due to properties of the respective loss functions (see Section 3.2). The progress in expectile regression models follows that on quantile regression, albeit at a slower pace. Perhaps this can be attributed to issues related to the interpretability of expectiles, as already explained in Section 3.2 (Waltrup et al. 2015).

Naturally, similar themes examined in quantile regression, find their counterparts in expectile regression. To mention some, one can find semiparametric models (Sobotka and Kneib 2012, Sobotka et al. 2013, Spiegel et al. 2017), regularization (Waldmann et al. 2017, Zhao and Zhang 2018, Zhao et al. 2018, Liao et al. 2019), Bayesian statistical modelling (Waldmann et al. 2017) and survival analysis (Seipp et al. 2021)

## 4.3   Forecasting with time series models

Let $\{y_1, ..., y_t\}$ be a time series indexed by $i = 1, ..., t, t \geq 1$. A time series forecasting problem is defined as the prediction of a future variable $\underline{y}_{t+h}$, at forecasting horizon $h > 0$, conditional on observations $\{y_i\}$, $i \leq t$. One-step ahead forecasting is defined for $h = 1$ and multi-step ahead forecasting is defined for $h > 1$. When the probability distribution of $\underline{y}_{t+h}$ is of interest, we have a probabilistic forecasting problem. The term automatic forecasting, that describes systems that forecast time series given only observed input and required in the case of large number of time series, extends also in the case of probabilistic forecasting (Ord and Lowe 1996).

Compared to IID data modelled by traditional regression models, time series data contain additional information due to temporal dependence; therefore, models may benefit by incorporating such information. A category of such models are stochastic processes, defined by $\{\underline{y}_1, ..., \underline{y}_t\}$. In the usual Bayesian setting, incorporating temporal dependence complicates things a little, in the sense that eq. (3) should be corrected to exploit such information. Practically, the one-step ahead forecasting problem is defined after appropriate modifications of eq. (3), by

$$p(\underline{y_{t+1}}|\, y_1, ..., y_t) = \int p(y_{t+1}|\boldsymbol{\theta}, y_1, ..., y_t)\, p(\boldsymbol{\theta}|y_1, ..., y_t)\, \mathrm{d}\boldsymbol{\theta} \tag{31}$$

The likelihood function, used to estimate the posterior distribution of $\boldsymbol{\theta}$, is also more complicated since the dependence between the variables of the stochastic process should be modelled.

### 4.3.1 Bayesian forecasting models

Bayesian statistical modelling constitutes a formal means for estimating the predictive distribution of future variables in time series forecasting problems. Due to the complicated nature of the parameters' likelihood function, it is not always possible to derive explicit forms of the predictive distribution using non-informative distributions, in contrary to regression problems (see e.g. Section 2.1). The formal theory for probabilistic forecasting of time series using Bayesian modelling and stochastic processes can be found in the overview by Geweke and Whiteman (2006). Some concepts and methods that are representative of what can be met in practice follow:

**Autoregressive moving average (ARMA) models**: Bayesian modelling of ARMA models (that also include the special cases of AR and MA models) using conjugate priors and simultaneously estimating the required number of parameters has been done by Monahan (1983).

**Autoregressive fractionally integrated moving average (ARFIMA) models**: Bayesian modelling of ARFIMA models using non-informative priors has been done by Pai and Ravishanker (1996), while Durham et al. (2019) propose solutions for large time series.

**Autoregressive with exogenous variables (ARX) models**: Multi-step ahead probabilistic forecasting for ARX models haw been studied by Liu (1994), who implemented a normal-gamma prior parameter distribution, and by Liu (1995) who implemented conjugate prior distributions in an ARX model with random exogenous variables.

**Vector models**: Vector autoregressive (VAR) models have been studied by Sims and Zha (1998) who used informative priors, and vector ARFIMA (VARFIMA) models have been studied by Ravishanker and Ray ([2002](#)). Bayesian modelling of global vector autoregressive models (B-GVAR) has been examined by Cuaresma et al. ([2016](#)).

**Exponential smoothing models**: Exponential smoothing models (including their variants, e.g. the Holt-Winters model) are statistical models specialized for forecasting. Some variants of exponential smoothing models may be subcases of ARMA models, but are referred as a different category, due to their success in practical applications. Bayesian settings for such models have been proposed, e.g. by Bermúdez et al. ([2010](#)) for the Holt-Winters model, and by Bermúdez et al. ([2009](#)) for the multivariate Holt-Winters model.

**Non-linear time series models**: AR models with time varying parameters (i.e. a class of non-linear time series model) has been examined in Bayesian settings by Müller et al. ([1997](#)).

**Generalized autoregressive conditional heteroscedasticity (GARCH) time series models**: Bayesian inference for GARCH time series models and its variant EGARCH (exponential GARCH) has been done by Vrontos et al. ([2000](#)). Bayesian inference methods for a variety of GARCH models, including e.g. the Glosten–Jagannathan–Runkle (GJR-ARCH) model have been reviewed by Virbickaite et al. ([2015](#)).

**Approximate Bayesian forecasting**: In time series forecasting problems, the size of the length of the time series may prohibit Bayesian methods from being practically applicable. As already noted in [Section 4.1](#), approximate Bayesian computation may be a benefitting solution with little cost in predictive performance (Frazier et al. [2019](#)).

### 4.3.2 Bootstrap-based forecasting models

Bayesian statistical models are parametric and may not be preferable in many cases; see related discussion in [Section 10](#). An alternative method to obtain probabilistic forecasts with time series models is resampling (or bootstrapping) techniques (Efron [1979](#)). The distribution of the innovations of the time series model (innovation is the random component of the model that essentially models its error) can be approximated by resampling the residuals of the fitted model. A first overview of such models can be found in Chatfield ([1993](#)). Some more recent resent developments include:

**Autoregressive moving average (ARMA) models**: The case of highly persistent AR models has been studied by Clements and Kim ([2007](#)).

**Vector models**: Bootstrap-based probabilistic forecasts have been studied by Grigoletto ([2005](#)). The case of multi-step ahead probabilistic forecasts has been investigating by Staszewska-Bystrova ([2011](#)), Fresoli et al. ([2015](#)) and Fresoli ([2022](#)).

**State-space models**: The case of periodic state-space models has been investigated by Guerbyenne and Hamdi ([2015](#)).

**Non-linear time series models**: Self-exciting threshold autoregressive (SETAR) model have been examined by Li ([2011](#)).

### 4.3.3 Quantile regression-based models

Quantile regression is an alternative to Bayesian statistical models for probabilistic predictions of regression models; see Section 4.2.1. Therefore, it is natural to expand such modelling techniques to time series models. Time series models, introduced in Sections 4.3.1 and 4.3.2, have their quantile loss based counterparts, as follows:

**Autoregressive moving average (ARMA) models**: AR models fitted with the quantile loss function have been investigated by Koenker and Xiao ([2006](#)). Quantile crossing issues have been addressed by Gourieroux and Jasiak ([2008](#)). A development related to the usual fitting procedure of ARMA models that considers correlations has been investigated in the context of quantile-based models by Li et al. ([2015](#)).

**Conditional Autoregressive Value at Risk (CAViaR) models:** CAViaR models predict quantiles by specifying their evolution over time using an autoregressive process (Engle and Manganelli [2004](#)).

**Conditional Autoregressive Expectile (CARE) models:** CARE models are similar to CAViaR models, with their difference being that they are estimated using an expectile loss function (Kuan et al. [2009](#)).

**Exponential smoothing models**: Quantile regression for exponential smoothing models has been studied by Taylor and Bunn ([1999](#)).

**Generalized autoregressive conditional heteroscedasticity (GARCH) time series models**: GARCH models have been investigated by Xiao and Koenker ([2009](#)).

### 4.3.4 Other forecasting models

Alternative practices for probabilistic forecasting with time series models are presented in Chatfield ([1993](#)) and include a Gaussian analytical approximation of the model's errors or approximation of the variance of the model's error. Progress in alternative techniques is also of interest, since relevant ideas may be applicable to general machine learning frameworks. Some examples include probabilistic forecasts of VAR models using computationally efficient algorithms for computing multi-dimensional integrals involving the Gaussian density function (Chan 1999), Gaussian approximations (De Luna 2000), first-order Taylor approximations (Snyder et al. 2001), approximations based on asymptotic properties (Hansen 2006) and approximation of the error's distribution (Wu 2012, Lee and Scholtes 2014). The problem of multi-step ahead probabilistic forecasting has been too complex to solve (Regnier 2018).

## 4.4 Copula-based regression models

Copulas are multivariate cumulative distribution functions with uniform marginal probability distributions for each variable (Nelsen 2006). The importance of copulas arises from Sklar's theorem (Sklar 1959), according to which, under certain conditions, there exists a unique copula function $C$, such that

$$H(x, y) = C(F(x), G(y)), \forall \, x, y \tag{32}$$

where $H$ is the joint distribution function of $x$ and $y$ with respective marginal distributions $F$ and $G$. The converse is true, i.e. if $C$ is a copula and $F$ and $G$ are distribution functions, then $H$ defined by eq. (32) is a joint distribution function with marginals $F$ and $G$.

Copula-based regression models have been introduced relatively very recently; however, due to their resemblance to simpler statistical models, we prefer to introduce them in this Section. In practice, copula-based regression models work by substituting $F(x)$ with the marginal distributions $\boldsymbol{F(x)} = (F_1(x_1), ..., F_p(x_p))$ of the predictor variables. The idea of copula-based regression is to obtain the distribution $G$, conditional on $\boldsymbol{x}$ and specified $C$ (that models the dependence between variables) (Noh et al. 2013). An early exploitation of the idea can be found in Sungur (2005). Copulas are parametric models; therefore, the key is to specify correctly the copula function. Since copulas are parametric models, extensions to probabilistic predictions for the regression models are natural. Furthermore, the dependence between variables is modelled separately from the marginal distributions, thus facilitating the modelling procedure. In the following, we will present several version based on combinations of components presented earlier.

**Estimation procedures**: Noh et al. (2013) consider a semiparametric approach for estimating the model. In particular, they propose parametric modelling of the copula function and non-parametric modelling of the marginal distributions. However, when the copula function is misspecified, the results may be largely unreliable, because the regression function is monotone in the predictor variables (Dette et al. 2014).

**Direct quantile estimation**: Quantile estimation by using directly the estimated conditional distribution functions has been examined by Kraus and Czado (2017) using vine copulas; see also Größer and Okhrin (2021) for the usefulness of using this class of copulas. Their method was extended to improve the conditional likelihood by Tepegjozova et al. (2022).

**Quantile regression**: Copula-based regression has been extended to the quantile regression case, both for IID data and for time series by Noh et al. (2015). The quantile regression case is also examined by Rémillard et al. (2017).

**Survival analysis with copula-based regression**: Copula-based regression for estimating quantiles in survival analysis has been investigated by De Backer et al. (2017).

**Regularization in copula-based regression**: High-dimensional Gaussian copula-based regression has been investigated by Tony Cai and Zhang (2018). He et al. (2018) propose transforming the variable selection problem to a multiple testing one, thus reducing the number of redundant variables due to the stopping rule.

**Bayesian statistical modelling**: Bayesian statistical modelling for copula-based regression models has been proposed by Klein and Kneib (2016), who also add non-linear predictors for the parameters of the model.

**Various copulas**: Among various types of copulas, some authors focus on a specific class, that may have some merits in particular situations. These include, for instance, Gaussian copula regression (Masarotto and Varin 2017, Zhao et al. 2020) and vine copulas for handling mixed (continuous and discrete) data, conditional heteroscedasticity, large-scale, nonlinear and non-Gaussian data (Chang and Joe 2019, Liu and Li 2022). Several other models are surveyed by Kolev and Paiva (2009).

**Missing data**: The case of missing observations has been investigated by Hamori et al. (2020) using a calibration estimation approach.

## 5. Machine learning algorithms

Now that fundamental concepts and simpler models have been introduced, it is time to continue with machine learning models for probabilistic predictions. A description of most machine learning models can be found in textbooks, such as those by Hastie et al. (2009), while we consider that the reader is familiar with machine learning algorithms (e.g. with random forests or boosting algorithms) in the following. Transforming a machine learning model to issue probabilistic predictions is possible by combining relevant fundamental concepts introduced earlier (see also Mukhopadhyay and Wang 2020, for a relative view that focuses on the connections between parametric and non-parametric models).

## 5.1   Gaussian process regression

A Gaussian process is a stochastic process, with any finite number of the process variables having a joint Gaussian distribution (Rasmussen 2004). Due to the multivariate Gaussian property, the predictive distribution can be estimated in Bayesian settings similarly to Gaussian models (e.g. linear models as shown in Section 4.1 or forecasting models as shown in Section 4.3.1). The form of the predictive distribution for various cases of prior distributions can be found, for example, in Rasmussen and Williams (2006) and Jankowiak et al. (2020) or in Girard et al. (2003) when the scope is time series forecasting and Corani et al. (2021) when the scope is automatic time series forecasting.

Probabilistic predictions with nonstationary Gaussian process models have been proposed by Liang and Lee (2019), while the relevant case of high-dimensionality has been investigated by Risser and Turek (2020). Heteroscedasticity has been modelled with Gaussian processes by Binois et al. (2018) and Binois and Gramacy (2021).

## 5.2   Generalized additive models for location, scale and shape (GAMLSS)

Beyond Bayesian statistical and quantile regression models, another category is that of generalized additive models for location, scale and shape (GAMLSS, Rigby and Stasinopoulos 2005). GAMLSS are an extension of the more restrictive generalized additive models (GAMS, Hastie and Tibshirani 1986) and model directly the parameters of the predictive distribution as functions of the predictor variables in regression settings; therefore, the full predictive distribution is estimated, unlike quantile regression models that estimate functionals of the predictive distribution. For this reason, GAMLSS have also been termed as distributional regression models (Umlauf and Kneib 2018). The models are general in the sense that any probability distribution can be modelled, thus moving beyond the Gaussian one, and including parameters beyond the mean and variance (e.g. the skewness). The parameters can be modelled by parametric, as well as additive non-parametric functions (e.g. smoothing splines, penalized splines, local regression smoothers) of the predictor variables, while random effects (e.g. spatial random effects) can also be included as well as combinations of components. The model is trained by penalized likelihood optimization (see also Section 3.4, for the connection of likelihood with the log score). Therefore, GAMLSS models need specification of the form of the dependent variable's probability distribution function, the link function and the estimation procedure (Henzi et al. 2021b). Software implementation of GAMLSS models can be found in Stasinopoulos and Rigby (2007). The combination of earlier ideas that led to the development of GAMLSS can be found in Green (2013). Various improvements and variants of GAMLSS models have also appeared (many are also summarized by Stasinopoulos et al. 2018). Those include:

**Alternative distributions**: Modelling of distributions with four parameters has been proposed by Rigby and Stasinopoulos (2006). Extensive investigation of symmetric probability distribution functions has been done by Ibacache-Pulgar et al. (2013).

**Copulas**: GAMLSS is extended to the case of predicting multiple responses modelled by copulas (Marra and Radice 2017) and to the case of predicting multiple instances of a single variable by modelling their dependencies using copulas (Smith and Klein 2021).

**Bayesian statistical modelling**: An extension of GAMLSS, termed Bayesian Additive Models for Location, Scale, and Shape (BAMLSS), that includes Bayesian statistical modelling has been proposed by Umlauf et al. (2018, 2021) and Umlauf and Kneib (2018). These authors bring Bayesian-based benefits related to efficient estimation procedure and statistical inference. Bayesian multivariate modelling (i.e. prediction of distributions of multiple variables simultaneously) has been investigated by Klein et al. (2015a). The Poisson model for an excess number of zeros has been especially investigated by Klein et al. (2015b) in the previous context. Extension to the case of skew distributions has been proposed by Michaelis et al. (2018).

**Other models**: The case of using instrumental variables in distributional regression has been investigated by Briseño Sanchez et al. (2020). Efficient estimation of censored or truncated dependent variables has been implemented by Messner et al. (2016).

## 5.3  Random forests

Random forests are ensembles of decision trees that are built using bootstrap aggregation with some additional randomization (Breiman 2001a). Quantile regression forests (Meinshausen 2006) are a generalization of random forests that can estimate conditional quantiles in a non-parametric way. Trees in quantile random forests are grown similarly to random forests and the conditional distribution is again estimated using the tree responses, although instead of the average of the ensembles, an indicator function is used to define quantiles. Quantile regression forests are consistent for quantile estimation under certain general conditions. Several other variants and improvements of quantile regression forests have been proposed, as presented in the following (see also the comparison of variants by Roy and Larocque 2020):

**Bias correction**: Quantile regression forests prediction bias is corrected in a post-processing framework in which the error of the predictions is subtracted at a second stage (Tung et al. 2014, Nguyen et al. 2015).

**Splitting rule**: In decision trees, the splitting rule (e.g. rules based on Gini impurity, variance reduction) is of great importance. Bhat et al. (2015) proposed a splitting rule based on the quantile loss function that resembles quantile regression. Generalized random forests are based on a gradient-based splitting scheme to optimize heterogeneity (Athey et al. 2019). Local linear forests model smooth signals and fix boundary bias issues (Friedberg et al. 2020).

**Distributional regression**: Distributional regression for random forests has been proposed by Schlosser et al. (2019) in similar way to GAMLSS, but using decision trees as base learners. An application on circular data using the von Mises distribution has been demonstrated by Lang et al. (2020). Modelling of the Beta distribution, which is particularly useful for variables bounded in (0, 1) has been proposed by Weinhold et al. (2020) using a tailored likelihood based splitting rule.

**Other models**: An online learning variant of quantile regression forests has been developed by Vasiloudis et al. (2019). Out-of-bag (OOB) prediction intervals have been proposed by Zhang et al. (2020). With this approach, a single predictive distribution is estimated, instead of separate conditional distributions for new cases. Random forests tailored for time series data (which have some additional useful properties related to temporal dependence; see Section 4.3) have been developed by Davis and Nielsen (2020) and may have potential for probabilistic prediction. The framework of Lu and Hardin (2021) for estimation of prediction intervals is based on OOB weighting of OOB prediction errors.

## 5.4   Boosting algorithms

Boosting algorithms are ensemble learning algorithms, in which the base learners are trained sequentially to minimize a loss function (Friedman 2001). Base learners are added in the ensemble with the aim to correct the errors of previous base learners. Two key components of boosting algorithms are the loss function that can be tailored to user needs and the type of base learners, in the sense that multiple types of base learners can be combined in an ensemble. Furthermore, boosting algorithms include an intrinsic mechanism for variable selection that is particularly suited for high-dimensional data (Mayr et al. 2014a, 2014b, 2017, Mayr and Hofner 2018). Friedman's (2001) original algorithm can optimize the quantile loss function, for example, by using decision trees as base learners. Therefore, it can predict conditional quantiles. Several other variants of boosting can give probabilistic predictions, based on its properties. In the following, we present some of them:

**Quantile regression**: Boosting for quantile regression has been proposed by Hofner et al. (2014) using algorithms particularly suited for high–dimensional data and including a diverse set of base learners ranging from linear effects to splines, spatial effects, random effects and more. Boosting algorithms that have shown exceptional performance in practical situations, with characteristic examples of those being the XGBoost (Chen and Guestrin 2016) and LightGBM (Ke et al. 2017), can also be optimized with the quantile loss function.

**Expectile regression**: Expectile regression with boosting has also implemented in the above framework by Hofner et al. (2014).

**Distributional regression**: Distributional regression with boosting using diverse base learners (ranging from linear effects to splines, spatial effects, random effects and more) has been proposed by Mayr et al. (2012), while the respective software implementation can be found in Hofner et al. (2016). The algorithm is termed gamboostLSS, for boosting generalized additive models for location, scale and shape. Modelling of the Beta distribution for bounded dependent variables has been investigated by Schmid et al. (2013). Stability selection for including the important predictors in the boosting ensemble has been proposed by Thomas et al. (2018), while a technique to deselect unimportant base learners has been proposed by Strömer et al. (2022). Natural Gradient Boosting (NGBoost) for probabilistic prediction (Duan et al. 2020) is another boosting-based distributional regression algorithm that has been implemented with multiple proper scoring rules.

**Other models**: Random gradient boosting (Yuan 2015) is an algorithm that combines random forests and boosting for exploiting advantages from both techniques for probabilistic prediction. Contrast trees can be used to find possible lack of fit for a model, while models can be improved by applying contrast boosting which can also provide probabilistic predictions (Friedman 2020). Probabilistic gradient boosting machine (PGBM) are used to estimate the mean and variance, which can then be used to sample from a specified distribution (Sprangers et al. 2021).

## 5.5 A list of other machine learning models

Beyond the main classes of models presented in Sections 5.1-5.4, other classes of models less frequently used are presented in the following.

### 5.5.1 Bayesian Additive Regression Trees (BART)

Bayesian Additive Regression Trees (BART) are a sum of decision trees in which a regularization prior is imposed on the parameters of the model (Chipman et al. 2010). BART are motivated by boosting algorithms, while they allow for probabilistic predictions. A variant of BART based on multiplicative regression trees allows modelling heteroscedasticity (Pratola et al. 2020)

### 5.5.2 Transformation models

Transformation models are practically obtained by inverting the regression equation. An extension of transformation models for estimating predictive distributions has been proposed by Hothorn et al. (2014). In contrast to GAMLSS, conditional transformation models rely on less assumptions (Hothorn et al. 2014). Investigation of maximum likelihood estimators for conditional transformation models has been done by Hothorn et al. (2018), while the relevant software implementation can be found in Hothorn (2020a). Boosting-based training of conditional transformation models has been extensively investigated by Hothorn (2020b), while decision trees motivated implementation has been proposed by Hothorn and Zeileis (2021). Autoregressive transformation models (that combine properties of time series models and transformation models) have been introduced by Rügamer et al. (2023a).

### 5.5.3 Conformal prediction

Conformal prediction is a technique to estimate marginal prediction intervals that include the dependent variable with a specified frequency (Vovk et al. 2005, Shafer and Vovk 2008, Gammerman 2012). The algorithm is based on a nonconformity measure (e.g. one that measures some kind of distance) for a given significance level and includes predictions conditional on the values of the measure using *p* values. The algorithm can be combined with any machine learning algorithm (e.g. those presented earlier). The method seems to be different from the example of proper scoring rules, while as also stated in Section 3.3.4, it is difficult to compare intervals with specified coverage probability.

**Conformal regression with various machine learning algorithms**: Conformal prediction with *k*-nearest neighbours regression (*k*-NN) combined with the introduction of six nonconformity measures has been investigated by Papadopoulos et al. (2011). Conformal prediction with *k*-NN for time series forecasting has been investigated by Tajmouati et al. (2022). Conformal prediction with random forests has been presented by Johansson et al. (2014), with survival random forests by Bostrom et al. (2017) and with Mondrian trees by Johansson et al. (2018).

**Conformal prediction with quantile regression algorithms**: Conformalized quantile regression combines conformal prediction with quantile regression. In particular, heteroscedasticity is modelled by quantile regression, while the coverage probability of the prediction interval is guaranteed. The method is applicable to all machine learning quantile regression algorithms (Romano et al. 2019, Sesia and Candès 2020).

**Conformal prediction in Bayesian settings**: Fong and Holmes (2021) combine the benefits of conformal prediction and Bayesian statistical modelling. Regularization may become a feature of the algorithm through using suitable prior distributions, the algorithm becomes more computationally feasible while keeping the coverage properties of conformal prediction.

**Other models**: Conformal prediction is marginal coverage in the sense that it is not conditioned on the predictor variable. An extension to conditional coverage is introduced by Lei and Wasserman (2014).

*5.5.4 Support vector regression.*

A semiparametric support vector regression variant of quantile regression has been proposed by Shim et al. (2012). Expectile regression with support vector machine regression has been proposed by Farooq and Steinwart (2017), while changing the quantile loss function to an orthogonal one makes support vector regression suitable for noisy data (Yu et al. 2018).

### 5.5.5 Splines

Quantile regression with splines has been investigated by He and Ng (1999). Non-crossing quantile regression has been explored by Kitahara et al. (2021) and by Xu and Reich (2021) in a Bayesian setting.

### 5.5.6 Local polynomial fitting

Polynomial quantile regression has been combined with quantile regression trees for probabilistic predictions by Chaudhuri and Loh (2002). Adam and Gijbels (2022) propose expectile regression with local polynomial fitting.

### 5.5.7 Functional regression

A general framework of regression models for functional data that includes quantile regression, as well as distributional regression, has been proposed by Greven and Scheipl (2017, see also Scheipl et al. 2016). Models are estimated using penalized likelihood or gradient boosting (Brockhaus et al. 2020). The model may include scalar or functional covariates of multiple types (e.g. linear effects, random effects, splines and more).

Regarding functional time series probabilistic forecasting, Hyndman and Shang (2009) propose bootstrap techniques. Moreover, probabilistic univariate time series forecasting has been addressed in a similar context by Shang and Hyndman (2011). Conformal prediction for multivariate functional data has been proposed by Diquigiovanni et al. (2022) and by Kelly and Chen (2022) for multiple functional regression.

### 5.5.8 Fuzzy logic

Methods for probabilistic forecasting with fuzzy time series has been proposed by Silva et al. (2016, 2020). They are based on specialized techniques that are used in time series forecasting and are not related to those presented in previous sections.

### 5.5.9 Converting deterministic to probabilistic predictions

The majority of environmental models issue point predictions, also known as deterministic predictions, in the field. To address this, some ideas have emerged to transform them into probabilistic predictions. These methods include the analog ensemble and the method of dressing, which have been formalized by Yang and van der Meer (2021) for forecasting applications. Given that point predictions are also issued in most machine learning applications, these ideas can be transferred to this field as well.

The analog method (Lorenz 1969) is based on the assumption that a current pattern of a time series is similar to past patterns. Therefore, what was realized in the past is likely to be realized in the future. This method uses strategies to find similar patterns in the past and use them to forecast the future. Applications of analog methods can be found in Alessandrini et al. (2015) and subsequent works. However, it is not always possible to identify similarities between past and current patterns when forecasts are issued by machine learning algorithms (Yang and van der Meer 2021). In this case, the method of dressing is preferred (Roulston and Smith 2003). This method estimates uncertainty by dressing past errors to current point forecasts.

### 5.5.10  Conditional probability density estimation with kernel smoothing

Kernel smoothing is a prominent non-parametric approach for conditional density estimation. In particular, the conditional probability distribution in a regression setting can be estimated by applying kernel smoothing to the empirical probability density distributions that are components of the conditional probability density function, i.e. (a) the joint distribution of the predictor and dependent variables and (b) the marginal probability distribution of the predictor variables (Hyndman et al. 1996). Kernel smoothing serves two scopes, i.e. (a) to smooth the empirical probability density distributions, while (b) simultaneously weight data points based on their similarity values (Gneiting and Katzfuss 2014). A review of related methods can be found in Rothfuss et al. (2019).

### 5.5.11  Other models

Other models for probabilistic predictions include the ensemble model output statistics (EMOS) by Gneiting et al. (2005). The method is a variant of linear regression that uses ensemble predictions as predictors and models the conditional mean and variance of a conditional Gaussian distribution. The method resembles distributional regression, but it further exploits properties that are met in weather forecasts when the latter take the form of ensembles. The mean and variance of a possibly non-Gaussian distribution has been modelled also by Nott (2006). The dependent variable is modelled by a double exponential family distribution and predictors are modelled by penalized splines, while inference is performed with a Bayesian framework. Again the proposed model resembles distributional regression. In a similar to GAM modelling context, Fasiolo et al. (2021) propose estimating directly the quantiles of conditional response distribution based on the quantile loss function in a Bayesian setting, thus avoiding to specify the probability distribution of the dependent variable as done in GAMLSS modelling.

Isotonic distributional regression is a case of distributional regression, in which the predictor space is equipped with a spatial order, while the conditional distribution is estimated using suitable loss functions in a way that the partial order is respected. The method exploits the partial order relationships. Compared to distributional regression models (see also previous sections), isotonic distributional regression is automatic after defining the spatial order. Frameworks for isotonic distributional regression are proposed by Henzi et al. (2021a, 2021b).

## 6. Neural networks and deep learning

Neural networks and deep learning models (i.e. those composed by multiple processing layers; Lecun et al. 2015, Schmidhuber 2015) including their variants (e.g. recurrent neural networks (RNNs), with the special case of long short-term memory (LSTMs), Hochreiter and Schmidhuber 1997, or convolutional neural networks, CNNs) are examined separately in this section due to their increasing significance. Previous reviews on probabilistic predictions have already identified several categories of neural networks and deep learning (see e.g. Khosravi et al. 2011, Kabir et al. 2018, Ovadia et al. 2019, Abdar et al. 2021). Here, we set the relevant material in the context developed previously in the paper.

### 6.1 Bayesian methods

Bayesian neural networks work similarly to Bayesian models defined earlier. Priors are assigned to model's parameters (weights on the connections and variance of the error term, Lee 2000) and posterior inference is possible for parameters, resulting in consistent posterior distributions (Lee 2000), while predictive distributions can also be estimated in a similar way. Reviews on Bayesian neural networks can be found in Lampinen and Vehtari (2001) and Titterington (2004). Some variants of Bayesian neural networks include those by Rohekar et al. (2019) who assigned prior distributions that depend on unlabelled predictors and Harva (2007), where the distribution of the dependent variable is a mixture of Gaussian distributions. The latter approach allows modelling the dependent variable with non-Gaussian distributions, thus being more flexible.

Due to heavy computational burden of Bayesian methods, variational inference methodologies have been developed to approximate the posterior distribution of the parameters (Swiatkowski et al. 2019). In particular, a probability distribution of a parametric form is specified and is estimated by minimizing the Kullback-Leibler (KL) between the specified distribution and the true posterior distribution. A likelihood-free variational inference methodology that allows for non-tractable variational densities, resulting in richer structures has been proposed by Tran et al. (2017).

## 6.2 Monte Carlo simulation-based methods

Since Bayesian techniques are computationally expensive, simulation techniques that can approximate Bayesian outputs may be efficient solutions when probabilistic predictions are of interest. Dropout (Srivastava et al. 2014) is one such technique in which during training, some layer outputs are dropped. Repeating training, one can obtain an ensemble of neural networks. The overall procedure is a regularization method that prevents overfitting. It has been proved (Gal and Ghahramani 2016) that dropout is an approximation of probabilistic deep Gaussian process in a Bayesian context, while it is computationally faster. In particular, predictive uncertainty with dropout can be estimated by calculating the variance of the ensemble.

Batch normalization is another Monte Carlo simulation-based technique that is used for regularization of neural networks. It has been proved that ensembles of neural networks obtained by batch normalization can again approximate uncertainty estimated in a Bayesian context (Teye et al. 2018). Another similar idea by Antorán et al. (2020) is to model the neural network depth as a random variable, resulting in Depth Uncertainty Networks (DUNs). In DUNs, marginalization over depth resembles Bayesian Model Averaging (BMA; see the explanation on combination methods in Section 8.2) over the ensemble of NNs (with each being deeper than previous ones). Another idea based on randomization comes from Mancini et al. (2021), who create ensembles of neural networks by randomizing the architectures (each neural network has different layer specific widths) using a Bernoulli mask. Prediction intervals are created by estimating the variance of the ensemble members.

## 6.3 Quantile and expectile regression

Probabilistic predictions using scoring functions is a natural choice for neural networks, since an integral part of their structure is exactly the objective function. Quantile regression neural networks have first been proposed by Taylor (2000) with software implementation by Cannon (2011). Xu et al. (2017) and Tagasovska and Lopez-Paz (2019) expand the quantile regression neural network to the case that the loss function is averaged at multiple quantile levels (this technique resembles the quantile scoring rule based on the quantile loss function, see Section 3.3.3). Modelling simultaneously at multiple quantile levels also mediates the quantile crossing problem, while also borrows information among various levels, thus producing more accurate predictions.

Related quantile regression neural networks variants include the recurrent neural network by Xie and Wen (2019), in which quantiles at multiple levels are estimated simultaneously. Quantile regression with tensors as predictors has been proposed by Lu et al. (2020). Tensor methods are implemented in deep learning models for improving estimation procedures. Moon et al. (2021) propose learning multiple quantiles with neural networks, by an optimization procedure that imposes an $L_1$ penalty to the gradient of the loss function. The resulting algorithm addresses the crossing-quantiles problem. In line with Bayesian quantile regression settings, Jantre et al. (2021) develop a similar one for neural networks that inherits merits of Bayesian models (see e.g. Bayesian statistical modelling and quantile regression in Section 4.2.1). Deep learning quantile regression for right censored data for has been proposed by Jia and Jeong (2022). Expectile regression with neural networks by implementing the expectile loss function has been proposed by Jiang et al. (2017). Huber quantile regression with neural networks has been proposed by Tyralis et al. (2023). Quantile regression neural networks and expectile regression neural networks are edge cases of Huber quantile neural networks.

## 6.4   Distributional regression

Distributional regression with neural networks is natural given the predictive abilities of neural networks. Early developments include Nix and Weigend (1994), in which the outputs of a neural networks are the mean and the variance. Similarly, the algorithm proposed by Cannon (2012) is used to estimate parameters of a target probability distribution using a neural network. In a related context, Lakshminarayanan et al. (2017) train a neural network using a log-likelihood function (recall its relationship with proper scoring rules) to estimate the mean and variance of Gaussian distribution (the mean and variance are outputs of the neural network), while Rasp and Lerch (2018) implement the CRPS for a Gaussian family of distributions. Klein et al. (2021) combined a copula regression model with a neural network that can model a richer family of distributions compared to the Gaussian focused approaches mentioned before. A special case of distributional regression (using the Von Mises distribution) with deep learning for modelling directional (circular) statistics has been proposed by Prokudin et al. (2018). A more general framework for distributional regression with deep learning has been proposed by Rügamer et al. (2023b, c), while transformation models in deep learning can be found in Kook et al. (2023).

## 6.5 Other neural networks models

Other variants for probabilistic predictions are mostly based on techniques already presented earlier, but tailored for neural networks. These include the following.

**Conformal prediction**: Conformal prediction with neural networks has been investigated by Demut and Holeňa (2012) for neural networks and Kompa et al. (2021) for deep learning. The special case of Recurrent Neural Networks (RNN) has been investigated by Stankevičiūte et al. (2021).

**Transformation models**: Neural network based transformation models have been investigated by Baumann et al. (2021).

**Generative adversarial networks**: Generative adversarial networks (GANs) are generative models; therefore, their extensions for probabilistic prediction is natural. GANs consist of two deep learning models, specifically of a generative network that generates a sample and a discriminative network that evaluates the sample (i.e. it predicts whether is real or fake in the sense that the sample belongs to the training set). The two networks compete in minimax two-player game regarding their losses (Goodfellow et al. 2014).

GANs are trained by minimizing a statistical divergence. Pacchiardi et al. (2022) proposed training a GAN using scoring rules, aiming to implement them for probabilistic forecasting. Conditional GANs are a formulation of GANS that allows conditioning a model on some data. Those techniques can be interpreted as regression problems in which GANs can provide probabilistic predictions and have been exploited by Koochali et al. (2019) in a probabilistic time series forecasting context and by Koochali et al. (2021) for multivariate probabilistic time series forecasting.

**Other variants**: Kuleshov et al. (2018) proposed an algorithm that focuses on calibration properties of the regression model, thereby deviating from the principle of maximization of the sharpness of the predictive distribution subject to calibration (see also Section 3.3.2; recall also the discussion of the specification of coverage probabilities in Section 3.3.3). They implement the algorithm with Bayesian deep learning models. Thiagarajan et al. (2020) proposed using different models for estimating the mean and prediction intervals of the dependent variable, and exploit estimates of prediction intervals for improving mean estimates in a bi-level optimization setting. An empirical assessment shows that their model produces good uncertainty estimates.

Li et al. (2021) propose an alternative solution, in which they partition the range of the dependent variable and use a multi-class classification model to assign probabilities to the predicted dependent variable depending on which bin it falls. The classification probabilities are then easily transformed to distributions. Probabilistic predictions in federated learning have been proposed by Thorgeirsson and Gauterin (2021), who assigned probabilities to the local weights of the model.

## 7.  A representative simple example

To better understand what probabilistic predictions are and how different models can estimate predictive distributions and be compared regarding the accuracy of their predictions we composed an example using simulated data. In particular, we simulated the following linear model:

$$\underline{y} = x_1 + x_2 + \underline{\varepsilon}, \ \underline{\varepsilon} \sim N(0, 1^2) \tag{33}$$

where $\underline{\varepsilon}$ is standard Gaussian noise. $\underline{x}_1$ and $\underline{x}_2$ were also simulated by a standard Gaussian distribution (recall from Section 2.1 that the choice of distribution is irrelevant to the result) and the resulting $y$ was obtained as the sum of eq. (33). We simulated 100 samples and used 50 samples for training and 50 samples for testing. Quantile regression and a GAMLSS with Gaussian dependent variable were fitted to the training set. Probabilistic predictions of the samples in the test set at multiple quantile levels with both algorithms is presented in Figure 2.

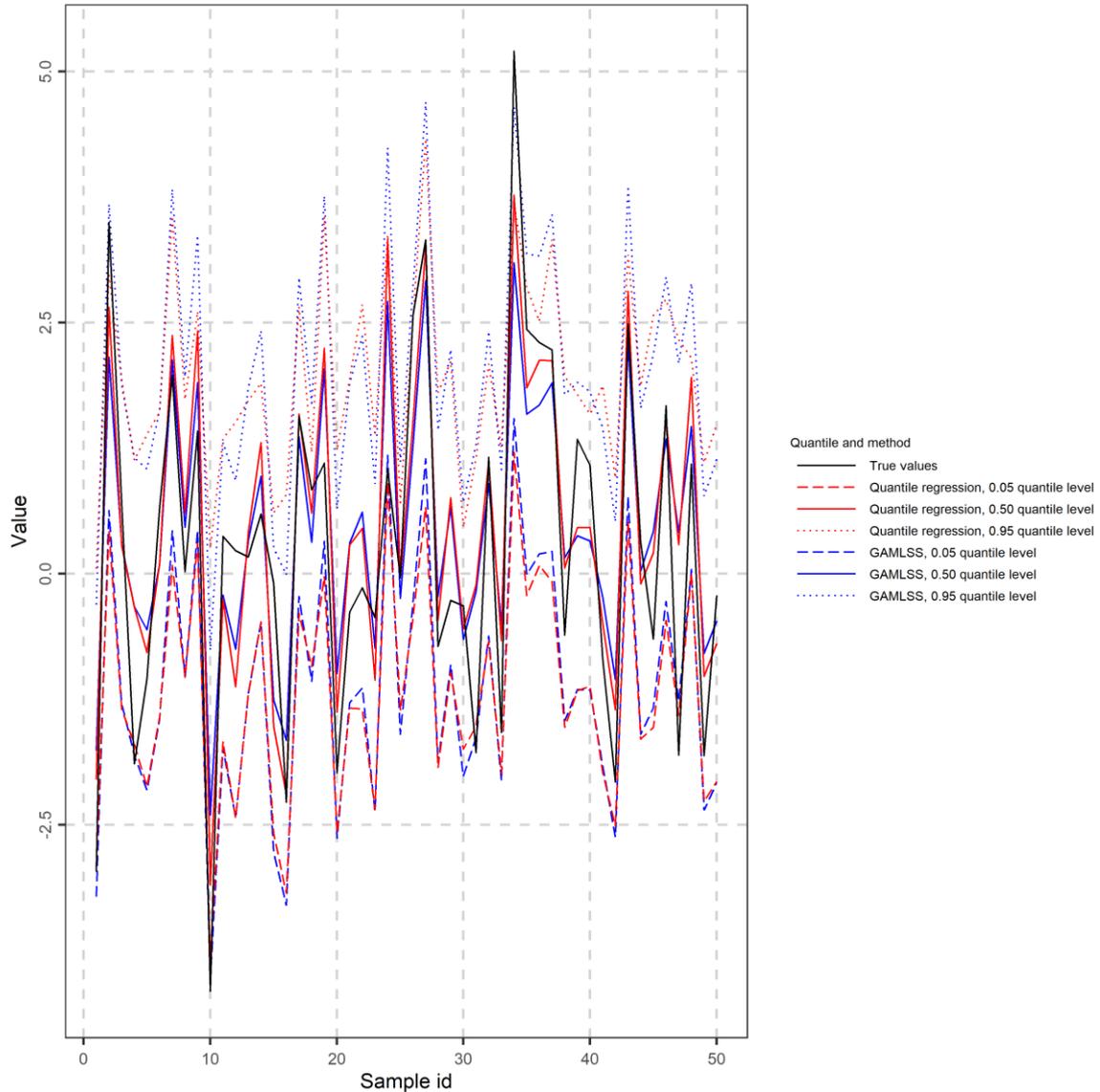

The simulation was repeated 10 000 times to have an accurate assessment of the performances of both methods using simulations of 100 (see e.g. the random case in Figure 2) and 1 000 samples (with 500 samples for training and 500 samples for testing). Coverage probabilities and quantile scores were averaged for the 10 000 cases, for both sizes of samples (100 and 1 000) and are presented in Table 2. Since quantile scores are not scale free, coverage probabilities show the difference in performance of both methods for different sample sizes (but recall that coverage probabilities are not proper scoring rules). Both methods become more accurate as the sample increases. For instance the mean coverage probabilities at level $\alpha = 0.05$ are 0.07344 and 0.05211 for sample sizes 100 and 1 000 for GAMLSS, respectively.

Table 2. Results of simulation experiments for quantile regression and GAMLSS.

| Sample size | Model | Coverage probabilities | | | Quantile scores | | |
| --- | --- | --- | --- | --- | --- | --- | --- |
| | | $\alpha = 0.05$ | $\alpha = 0.50$ | $\alpha = 0.95$ | $\alpha = 0.05$ | $\alpha = 0.50$ | $\alpha = 0.95$ |
| 100 | Quantile regression | 0.07921 | 0.49943 | 0.92112 | 0.12044 | 0.41847 | 0.11988 |
| 100 | GAMLSS | 0.07344 | 0.49890 | 0.92677 | 0.11460 | 0.41317 | 0.11426 |
| 1 000 | Quantile regression | 0.05281 | 0.50034 | 0.94728 | 0.10456 | 0.40095 | 0.10455 |
| 1 000 | GAMLSS | 0.05211 | 0.50026 | 0.94793 | 0.10391 | 0.40027 | 0.10389 |

Regarding quantile scores, GAMLSS performs better compared to quantile regression at all quantile levels and all sample sizes. However, the difference between the two methods decreases as the sample size increases. One should expect that GAMLSS would be better because the distribution of the depended variable is well specified (recall that this is a simulation experiment where properties of variables are known a priori), thus giving an advantage to parametric models. As the sample size increases, non-parametric models improve at a higher rate but still remain worse compared to the parametric ones. The situation could be reversed in real data examples where the chance of a misspecified probability distribution of the dependent variable becomes high; therefore, non-parametric methods start to prevail.

## 8. Combinations of algorithms

Combinations of algorithms, also termed ensemble learning in the machine learning field have been proved to improve over the individual algorithms with respect to multiple aspects, including predictive performance for point predictions. The same can be said for the combination of algorithms that issue probabilistic predictions. That direction of research of probabilistic forecasting has been particularly developed by the forecasting community, following the vast knowledge on the combinations of algorithms that deliver point forecasts. An overall view of the field can be found in Winkler et al. (2019).

### 8.1 Weighted averaging of algorithms

The simplest combinations are those that are based on weighted averaging of predictions. Now there are some different directions that can be followed. In the first crossroad, there is the dilemma of combining distributions or functionals of distributions (e.g. quantiles), with the former being met more frequently. After deciding on what to combine, a method to estimate the weights of the combination is needed. Combinations of algorithms may include, for instance, the weighted linear averaging (the "*linear opinion pool*") of cumulative distributions (eq. (34)) or densities (eq. (35)), or the multiplicative combination (the "*logarithmic opinion pool*") in eq (36); see e.g. the overview by Genest and Zidek (1986).

$$F = \sum_{i=1}^{K} w_i F_i \tag{34}$$

$$f = \sum_{i=1}^{K} w_i f_i \tag{35}$$

$$f = \prod_{i=1}^{K} f_i^{w_i} \,/\, \int \prod_{i=1}^{K} f_i^{w_i} \, \mathrm{d}\mu \tag{36}$$

In the right-above equations, $K$ is the number of algorithms to be combined, and $F_i$ and $f_i$ are the cumulative distributions functions (CDFs) and probability density functions (PDFs) predicted by algorithm $i$, respectively. Several other combination strategies also exist, including the centered linear pool (Knüppel and Krüger 2022). The expansion to the combination of quantiles or other functionals (e.g. expectiles) is straightforward.

The equally weighted linear pool (simple averaging) is a combination that is hard to beat in practice (Lichtendahl et al. 2013), with multiple attempts being made and being more successful as the data size increases (see e.g. the comparison by Clements and Harvey 2011 and the theoretical comparison by Gneiting and Ranjan 2013).

Regarding the choice of averaging distributions or quantiles when examining equally weighted linear pools, Lichtendahl et al. (2013) support the latter option, which is shown that it assigns sharper forecasts. Furthermore, Lichtendahl et al. (2013) have shown that the equal weight averaging performs no worse compared to the average of the scores of the individual algorithms. This finding is called "*harnessing the wisdom of the crowd*".

Random forests and boosting are ensemble learning algorithms. A single type of base learner is used (usually decision trees) by both algorithms. In Sections 5.3 and 5.4, we presented how random forests and boosting can assign probabilistic predictions. Stacking needs diverse types of algorithms and resembles the culture of forecasters in combining algorithms. We shall talk about stacking in Section 8.3, while earlier in Section 8.2 we shall talk Bayesian Model Averaging (BMA), which seems to be another standard practice of forecasters in combining probabilistic forecasts.

**Combining distributions**: Minimizing a loss function with respect to the weights of the implemented opinion pool is a usual method for estimating optimal weights. Hall and Mitchell (2007) proposed estimating the weights of the linear pool by minimizing a distance, measured by the Kullback–Leibler information criterion (which corresponds to the log-score), while Hora and Kardeş (2015) proposed estimation by minimizing the quadratic scoring rule. Opschoor et al. (2017) and Berrisch and Ziel (2021) estimated weights by minimizing the censored likelihood scoring rule as well as the CRPS. Ranjan and Gneiting (2010) recommended a beta-transform of the linear opinion pool and they estimated the parameters of the beta-transform and the weights by maximizing the likelihood of the model (recall the relationship between maximum likelihood and the logarithmic score in Section 3.4). The model is shown to improve over the linear opinion pool with regards to calibration and sharpness.

A more flexible option to allow estimated weights depending on the forecasted value, thus not being constant has been proposed by Kapetanios et al. (2015). Bayesian modelling of beta-transformed opinion pools has been proposed by Casarin et al. (2016) and Bassetti et al. (2018) resulting in better performances. Previous combinations refer to multiple predictions for a single point. Ray et al. (2017) proposed combining probabilistic time series forecasts at multiple horizons using copulas.

**Combining quantiles**: Estimation of quantile forecasts by minimizing a quantile loss function with respect to the weights has been proposed by Taylor and Bunn (1998), i.e. to apply quantile regression to linear opinion pool. Since linear in parameters quantile regression allows for a constant term in addition to the weights of the covariates, that constant term is also possible to be added in the linear pool. Taylor and Bunn (1998) have shown that, when quantile predictions are unbiased, the constant should be omitted and the weights should sum to unity. In Shan and Yang (2009), the weights are estimated by a ratio of multiplicative quantile losses (after being multiplied with a tuning parameter and transformed with an exponential function) divided by the loss of all algorithms.

Methods for the combination of prediction intervals have been investigated by Gaba et al. (2017) that may also apply to the combination of quantile forecasts and vice versa, since prediction intervals are defined by their upper and lower predictive quantiles. Besides the simple average, such methods include the median, the envelope of the interval, exterior as well as interior trimming and more.

## 8.2   Bayesian model averaging

Besides averaging algorithms, a natural alternative for combining models is in Bayesian settings, in particular Bayesian Model Averaging (BMA) (for a review on BMA, see Hoeting et al. 1999 and Kaplan 2021). Assuming that models $M_i$, $i = 1, ..., K$ are combined and $\underline{z}$ is the (future) variable of interest (recall the notation from Section 2.1), then the predictive distribution is given by

$$p(z|y) = \sum_{i=1}^{K} p(z|M_i, y)\, p(M_i|y) \tag{37}$$

where $p(M_i|y)$ is given by

$$p(M_i|y) = (p(y|M_i)\,p(M_i)) / (\textstyle\sum_{i=1}^{K} p(y|M_i)\,p(M_i)) \qquad (38)$$

and $p(y|M_i)$ is given by

$$p(y|M_i) = \int p(y|\theta_i, M_i)p(\theta_i|M_i)\mathrm{d}\theta_i \qquad (39)$$

in the usual Bayesian way.

Applications of BMA include improvements of forecasts ensembles that tend to be underdispersive (Raftery et al. 2005). Several options for assigning distributions to the ensemble members are possible, depending on the problem at hand. For instance, Baran (2014) implemented truncated normal distributions. Other applications include imputation of missing data (Kaplan and Yavuz 2020). BMA has also been used as an inherent component of machine learning models. For instance, it was applied to BART (Hernández et al. 2018; see Section 5.5.1) to obtain an algorithm that performs better when the number of predictors is large.

## 8.3   Stacked generalization

Stacked generalization (also called stacking; Wolpert 1992) is an ensemble learning algorithm that gains strength when combining machine learning algorithms with diverse properties. In contrast to combinations algorithms presented in Section 8.1, in which weights are estimated based on in-sample performance, stacking weights are estimated by out-of-sample performances. Implemented losses can be consistent scoring functions or proper scoring rules as those presented in Section 3.

Yao et al. (2018) favour the use of stacking over BMA for probabilistic predictions in cases that the set of combined algorithms does not include the true data generating model. The combiner algorithm could be linear (e.g. the weighted average in Section 8.1), but more flexible algorithms (e.g. quantile regression forests) could also be used for increasing performance.

## 9.   Special cases

Existing machine learning algorithms are adapted for specific applications for which some prior knowledge exists regarding properties of the problem under investigation. For instance time series forecasting problems are characterized by temporal dependence structures and spatial prediction problems are characterized by spatial dependence structures. Exploiting information from those dependence structures may result in improved predictive performances. Specialized algorithms for temporal, spatial and spatio-temporal models are presented in Sections 9.1 - 9.3.

Other specialized applications of probabilistic predictions that attract interest due to being especially relevant in scientific fields are also presented in the following. Those include, for example, predictions of extremes and uncertainty of measurements (see Sections 9.4 and 9.5, respectively). We conclude with applications of probabilistic predictions in various scientific fields (see Section 9.6).

## 9.1 Temporal models

A treatment of Gaussian processes for time series forecasting including the case of probabilistic forecasts can be found in Roberts et al. (2013). Modelling time series with an algorithm that borrows ideas from GAMs can be found in Taylor and Letham (2018). The proposed algorithm models various time series components, including seasonality, trends and effects of holidays, while it also assigns probabilistic forecasts.

Other models are mostly deep learning based with special features, following a recent explosion of deep learning use in forecasting competitions, although tree-based methods were also successful in those competitions (Januschowski et al. 2021). In fact, most deep learning algorithms are based on RNN architectures (Hewamalage et al. 2021), albeit exceptions also exist. Deep learning algorithms for probabilistic forecasting include:

**Hybrid models**: Those models integrate two or more algorithms and exploit properties from both algorithms. For instance, Khosravi et al. (2013) use a neural network to predict the mean of a time series and, subsequently, the predictions are transformed to probabilistic by imposing a GARCH model on top of the neural network to model the variance. Gasthaus et al. (2020) propose modelling the distribution of the response variable using spline quantile functions on top of a RNN, instead of modelling quantiles separately at each level. Smyl (2020) proposed a model that combines RNNs and exponential smoothing. The role of the latter algorithm in the model was to deseasonalize and normalize the time series.

**Quantile regression models**: Quantile autoregression neural networks that include neural networks combined with quantile autoregression have been proposed by Xu et al. (2016). A quantile regression neural network for mixed sampling frequency data has been proposed by Xu et al. (2021).

**Monte Carlo-based methods**: Implementation of dropout in LSTM (a type of RNN that is particularly resistant to optimization misbehaviours) for probabilistic forecasting has been proposed by Serpell et al. (2019). Bootstrap-based predictions intervals were forecasted by a neural network by Mathonsi and Van Zyl (2020).

**Distributional regression**: The method by Hu et al. (2020), in which the parameters of a variable bounded in [0, 1] is estimated by neural networks can be said to belong to the class of distributional regression. Variables with larger support can be transformed to ones bounded in [0, 1].

**Modelling multiple time series simultaneously**: Modelling multiple time series with a single model can exploit additional information compared to the case that each series is modelled separately and leads to large improvements regarding predictive performance (Sen et al. 2019). Examples include Chen et al. (2020), who implement a CNN architecture and allow for quantile and distributional regression, and Salinas et al. (2020), who implement an autoregressive component, while their model is of the distributional regression type.

**Software**: Software that unifies multiple models for probabilistic time series forecasting include Gluon Time Series Toolkit (GluonTS) by Alexandrov et al. (2020).

## 9.2   Spatial models

As already stated previously, dependence structures in spatial problems can be exploited to improve predictions similarly to time series models. However, compared to the usual case in time series modelling, an additional obstacle in spatial modelling is that measurements are placed in irregular locations. Machine learning models are also used for spatial predictions, among which Gaussian process regression holds a prominent place due to tradition in the field. Pure machine learning models are also applicable, although information from spatial dependencies is not exploited in a straightforward way (Hengl et al. 2018).

**Gaussian process regression**:  A Bayesian statistical model for Gaussian process modelling has been developed by Banerjee et al. (2008). The model is particularly suited for big datasets, while also remedying relevant weaknesses that are inherent characteristics of Bayesian models. Further developments to address issues of computational complexity with computationally efficient modelling have been approximate Bayesian inference by Eidsvik et al. (2012), convenient prior distributions (Duan et al. 2017) and nearest-neighbour Gaussian processes (Zhang et al. 2019).

Inserting Gaussian processes to a multi-layer model is another approach that exploits advances in deep learning (Zammit-Mangion et al. 2021).

**Markov random fields**: Markov random fields can be used to approximate spatial processes and are more appropriate for measurements in a regular grid (Banerjee et al. 2008). For a treatment of Gaussian Markov random fields, see Rue and Held (2005) and MacNab (2018). A GAMLSS approach to Gaussian Markov random fields has been proposed by De Bastiani et al. (2018).

**Quantile regression**: Quantile regression has been extended to the spatial case by Hallin et al. (2009). The spatial dependence has been modelled by copulas in a Bayesian statistical-based setting (Chen and Tokdar 2021).

**Additive models**: Bayesian modelling for additive models that may also include Markov random fields and spatial effects has been examined by Fahrmeir and Kneib (2009).

**Bayesian model averaging**: Bayesian modelling averaging adapted for spatial settings (termed geostatistical model averaging) has been proposed by Kleiber et al. (2011).

**Deep learning**: Sidén and Lindsten (2020) showed a connection between Gaussian Markov random fields and CNNs that allows generalization to multi-layer architectures (termed deep Gaussian Markov random fields).

## 9.3 Spatio-temporal models

Modelling of temporal and spatial data (Hefley et al. 2017) poses new challenges, while new more complex models are needed to address related problems. Those models have been particularly developed by the community of the spatial statistics field, and are mostly based on Bayesian statistics, with characteristic examples being available in Katzfuss and Cressie (2012) and Finley et al. (2015). Gaussian processes have also been used (Wang and Gelfand 2014). Expectile-based models (Spiegel et al. 2020) and deep learning (Wu et al. 2021) have also become options to model spatio-temporal data.

## 9.4 Extremes

Predicting events that happen infrequently is a major challenge for machine learning. Such events can be various disasters (e.g. floods) and unexpected events (e.g. extremely high demand for electricity that cannot be supported by the grid). When prediction of a variable is probabilistic and the interest is on modelling extremes, then the tails of the distribution or extremely low or high quantiles of the distribution should be modelled. Distributions that are suitable for such variables are usually heavy-tailed, i.e. a larger part of the mass is located at the tails of the distribution compared to light-tailed distributions (e.g. the Gaussian distribution). Estimation of the parameters in distributional regression approaches may be unstable at such cases, while quantile regression (or consistent scoring functions-based methods) may not suit for such problems, since conventional large-sample theory does not apply sufficiently far in the tails (Chernozhukov 2005).

In the following, we will discuss issues related to scoring functions for extremes and improvements on quantile regression for modelling extremes:

**Quantile regression**: Large sample properties of extremal quantile regression have been studied by Chernozhukov (2005). Wang and Li (2013) and Wang et al. (2012) proposed extrapolating intermediate quantiles, predicted with quantile regression, far in the tails using extreme value theory. Firpo et al. (2021) applied a generalized method of moments (GMM) estimation of quantile regression coefficients by using flexible parametric restrictions that allow to extrapolate intermediate quantiles.

**Expectile regression**: Extrapolation of intermediate expectiles far in the tails has been proposed by Daouia et al. (2018), while functional estimation has been proposed by Girard et al. (2022a). Furthermore, a non-parametric method for predicting extreme expectiles has been developed by Girard et al. (2022b) based on their distributional relationship with quantiles. Joint inference for several extreme expectiles has been investigated by Padoan and Stupfler (2022).

**Scoring extremes**: Diks et al. (2011) proposed using as scoring rules, a conditional likelihood, given that the event lies in the region of interest (in the tail of the distribution) or a censored likelihood, with censoring of events outside the region. Those scoring rules are more appropriate when the interest in on predicting the distribution at the specified region. Closed form expressions of the CRPS for the (heavy-tailed) generalized extreme value distribution (GEV) and the generalized Pareto distribution (GPD) were developed by Friederichs and Thorarinsdottir (2012). Weighted scoring rules with emphasis on the tails have been proposed by Lerch et al. (2017). The exceedance probability scoring rule, which is suitable to assess exceedance probabilities, was developed by Juutilainen et al. (2012), while Taggart (2022a) proposed consistent scoring functions for comparing point predictions in the region of tail of the distribution. Non-parametric scoring methods that are based on a cross-validation setting were also proposed by Gandy et al. (2022). However, expected scores are not appropriate to distinguish tail properties (Brehmer and Strokorb 2019).

**Other methods**: Oesting et al. (2017) proposed Brown-Resnick stochastic processes for forecasting extreme events. Brown-Resnick processes are max-stable; thus, they can model heavy tails, while they are also particularly suitable to model spatial random fields with max-stability properties.

**Deep learning**: Deep learning (in particular CNNs) has been used to estimate the type of dependence (specifically, to guess between asymptotic dependence or independence) of spatial extremes (Ahmed et al. 2021).

## 9.5   Uncertainty in measurements (observational errors)

Measurement errors are common in many fields (Hariri et al. 2019). For instance, measured quantities of rainfall over an area tend to be inexact since stations cannot cover the full region. Regression models are tailored to predict based on data with errors; thus, they do not report the real values. For this reason, adaptations of regression models are needed to account for these errors. Methods that account for several types of measurement errors have been developed when probabilistic predictions are needed. Some representative methods will be presented in the following.

Wei and Carroll (2009) developed a quantile regression method that corrects measurements errors in the predictors by adapting the quantile-loss based minimization problem, introducing a probability distribution to model the error. The case of measurement errors of the dependent variable has been examined by Ferro (2017) and Bessac and Naveau (2021), who adapted proper scoring rules to address those errors (see also Brehmer and Gneiting 2020, for a related example on proper scoring rules).

## 9.6   Some representative applications

The requirement for delivering probabilistic predictions has been recognized in several scientific fields. Consequently, review articles on aspects of methods for probabilistic predictions are met frequently. In the following, we discuss them.

### 9.6.1 Economics and finance

In economics and finance, Duran (2008) discusses methods for probabilistic forecasting of sales. Deep learning for financial time series forecasting has been surveyed by Sezer et al. (2020). Although, most applications regard point forecasts, the paper is a good starting point for probabilistic time series forecasting given the straightforward extensions to probabilistic deep learning, as already discussed earlier here. Finally, combination of algorithms in economics has been reviewed by Steel (2020).

### 9.6.2 Energy research

Applications of probabilistic predictions are frequent. They are summarized by papers that are discussed in the following.

**Renewable energy**: Probabilistic wind power generation forecasting has been reviewed by Zhang et al. (2014). Yang et al. (2022) reviewed solar forecasting including probabilistic methods, while van der Meer et al. (2018) surveyed probabilistic photovoltaic power production forecasting. Post-processing in solar forecasting has been reviewed by Yang and van der Meer (2021). Post-processing refers to correcting point forecasts (usually assigned by numerical weather models) and transforming them to probabilistic. Deep learning for wind and solar energy forecasting has been surveyed by Alkhayat and Mehmood (2021).

**Electricity price**: Electricity price forecasting, including methods for probabilistic forecasting, has been reviewed by Weron (2014). Nowotarski and Weron (2018) and Ziel and Steinert (2018) focused on methods for probabilistic electricity price forecasting.

**Electric load and electricity consumption**: Probabilistic electric load forecasting has been surveyed by Hong and Fan (2016). Van der Meer et al. (2018) surveyed probabilistic electricity consumption forecasting.

**Smart energy systems**: Probabilistic forecasting in smart grids has been surveyed by Khajeh and Laaksonen (2022) and Ahmad et al. (2022).

**Other topics**: Gensler et al. (2018) focus on directives for selecting metrics for probabilistic forecasting in renewable energy. The class of boosting algorithms that are especially relevant for probabilistic predictions in energy has been surveyed by Tyralis and Papacharalampous (2021).

### 9.6.3 Environmental and earth and planetary sciences

Applications of probabilistic predictions and forecasts are frequent in environmental, as well as in earth and planetary sciences, with machine learning playing an important role in those applications (Haupt et al. 2022). Beyond probabilistic predictions, related topics of interest include the prediction of extremes (Ghil et al. 2011, Sillmann et al. 2017, Huser 2021) and post-processing methods, with the latter topic appearing due to the need of integrating weather forecasting models, which are especially relevant to the field. We discuss them in the following.

**Post-processing methods**: Post-processing methods for weather and hydrological forecasting are reviewed by Li et al. (2017) and Vannitsem et al. (2021).

**Other topics**: Methods for probabilistic predictions in geology are surveyed by Albarello and D'Amico (2015). Probabilistic forecasting with machine learning in hydrology is reviewed by Papacharalampous and Tyralis (2022). Metrics for probabilistic predictions are surveyed by Huang and Zhao (2022) in hydroclimatology. The class of random forests algorithms that can also assign probabilistic predictions has been surveyed in hydrology by Tyralis et al. (2019b).

### 9.6.4 Big data comparisons

Comparison of multiple machine learning methods using big datasets is essential to understand their performance and allows to understand their properties. Such comparisons exist in the literature, although they are rare compared to comparisons of machine learning methods for point predictions (for the latter, see e.g. the survey by Fernández-Delgado et al. 2019). Such studies include the comparison of multiple algorithms in environmental data (Papacharalampous et al. 2019, Tyralis et al. 2019a), computer science problems (Torossian et al. 2020), data competitions (Grushka-Cockayne and Jose 2020) and spatial problems (Fouedjio and Klump 2019). Such data comparisons could be facilitated in the future by the implementation of specialized software (e.g. Ghosh et al. 2022).

## 10. Summary and future outlook

## 10.1 Setting a probabilistic prediction problem – technical considerations

Setting a probabilistic prediction problem requires understanding of all nuances in the field. That is possible by knowing the evolution in the field from the simplest statistical algorithms to more complex ones. As it was shown previously in this review paper, the foundation of the field lies on simple concepts developed at different time periods.

The first type of foundational concepts originate from the theory of Bayesian statistical modelling. Within this framework, parametric models can estimate the predictive distribution of the dependent variable, given that the model is correctly specified. Another foundational area lies in linear regression modelling. Linear regression models are among the simpler models and can be transformed to Bayesian statistical models to subsequently predict probabilities for future dependent variables. The third foundational area is consisted of scoring rules for training – calibrating statistical models. It allows: (a) fitting parametric models, thereby avoiding Bayesian-based simulations; and (b) fitting non-parametric models, thereby bypassing the problem of the possible misspecification of probability distributions. Based on the three above-outlined foundational areas, as well as on the theory of machine learning models, it is possible to build new models tailored to the problem at hand. Specific problems are already discussed in previous sections of this review paper, while a synthesis of seemingly unrelated methodologies follows. Those methodologies can serve as components of new algorithms for probabilistic predictions in ways that are clarified later in this section.

There are three main types of practical problems depending on the readily available information (see Figure 3). In the first one, the response variables are independent given the predictor variables. Most machine learning models are suited for this case. The case of temporal dependence (appearing when modelling time series) is well dealt by the time series models, as well as by specialized deep learning models (e.g. LSTMs) that focus on the exploitation of temporal dependence information. Obviously, the usual machine leaning models can also be applied to these cases, with occasionally great performance. A reason for their possible great performance in modelling time series data, despite their inability to exploit the information of temporal dependence, could be that their additional complexity and flexibility compensates for the lost information. Problems with spatial data, in which spatial dependence information might be exploited are usually addressed with spatial statistical models (e.g. kriging) and Gaussian processes. The latter models are preferred compared to other machine learning models in the field of spatial modelling perhaps due to reasons of tradition, although other machine learning models may also be applicable. Regarding the predictive performance in modelling in spatial settings using usual machine learning algorithms, similar arguments with the case of modelling time series using these same algorithms apply.

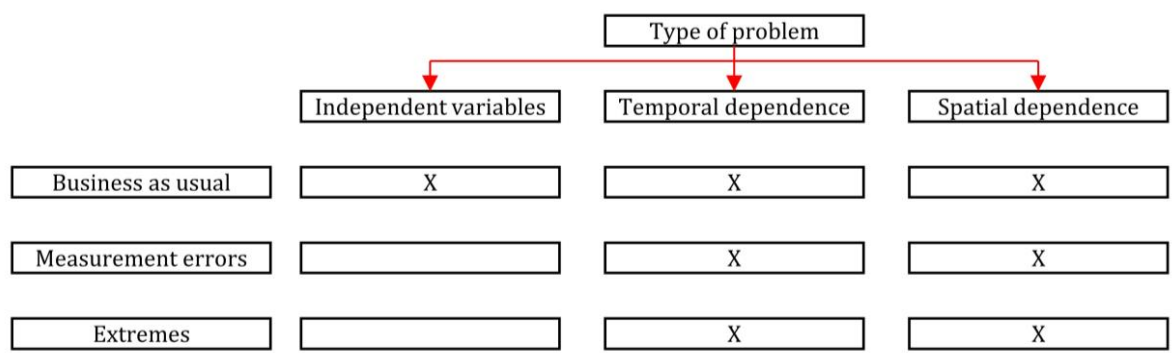

| | Independent variables | Temporal dependence | Spatial dependence |
|---|---|---|---|
| Business as usual | X | X | X |
| Measurement errors | | X | X |
| Extremes | | X | X |

Figure 3. Frequent types of probabilistic prediction problems and cases of special interest met at each of them.

"Business as usual" modelling is frequently met in all the three aforementioned types of practical problems. On the contrary, modelling of measurement errors and modelling of extremes are mostly required in time series and spatial settings. Although, the latter two modelling cases might also be observed in practical problems with independent variables, these occurrences seem to be relatively rare. Perhaps the main reason behind this fact, for the case of measurement error models, is that some types of observations are characterized by minimal error. This might indeed hold, for instance, in house pricing modelling, where modelling of independent dependent variables is effective. Modelling of extremes is met in environmental and finance applications, in which temporal or spatial dependence seems to be the norm.

Beyond the type of problem, the type of prediction is also of interest. For instance one might be interested in a specified functional of the predictive distributions. That might be, for example, the case in which one is interested on the probability that temperature exceeds a pre-specified quantity. The full predictive distribution of the dependent variable will be more informative, although its estimation might be harder. Those two problem definitions are presented in Figure 4.

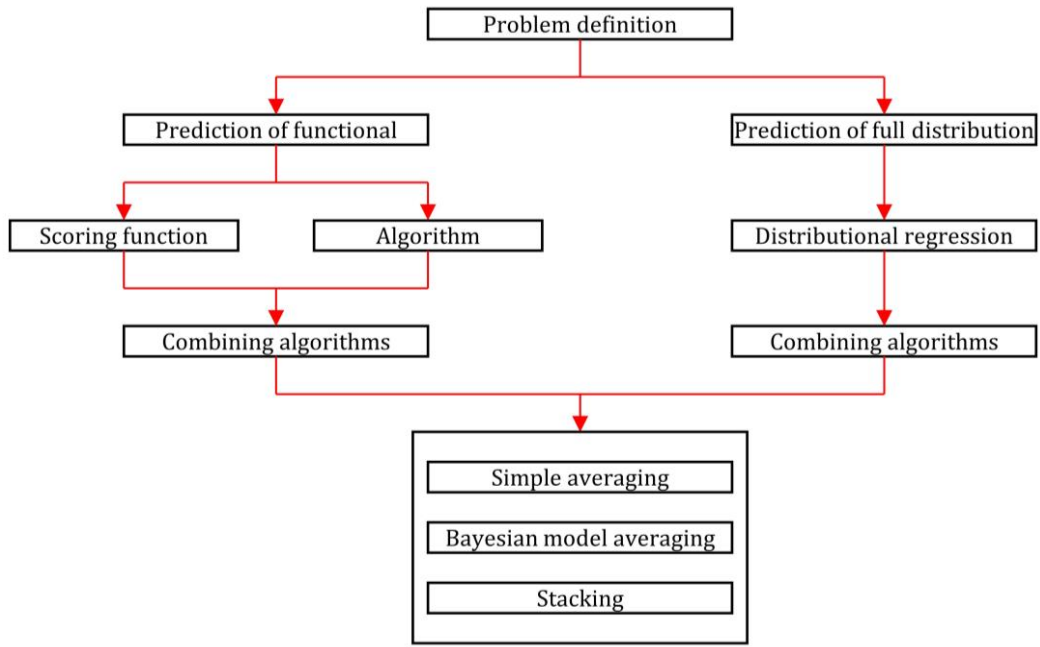

. Problem definition based on type of prediction and related ways for handling such problems.

When one is interested in predicting a functional, he/she should apply a consistent scoring function for this functional to fit a related machine learning model. For instance, a quick way to predict a quantile at a specific level is to fit a linear model using the quantile loss function. If one is interested in increased performance, then many algorithms may be fitted and then combined. The primary combiners are simple averaging, Bayesian model averaging and stacking, with each method having its merits and deficiencies. Similarly, distributional regression is the primary way to obtain the full predictive distribution, while fitting can be done with proper scoring rules. Again, the combination of algorithms is possible for obtaining improved predictive performance. Obviously, by estimating the full predictive distribution, functionals can be made readily available, while by estimating multiple functionals (e.g. quantiles at multiple levels) one can also get the predictive distribution at a high resolution. Both approaches have their merits, which are discussed in Section 10.2.

Figure 5 is representative on how elements from the three foundational areas can be combined to have a new algorithm that can be suited for a specific task. For instance, one might be interested in estimating the full predictive distribution using boosting. In this case, the most suitable algorithm might be the distributional boosting one described in Section 5.4. This algorithm is based on the optimization of a proper scoring rule applied to a machine learning algorithm. Another example is that of linear Bayesian statistical modelling, which is founded on the application of Bayesian modelling to a statistical learning algorithm (specifically, to the linear regression model), as described in Section 4.1. The expansion of this approach for dealing with the case of time series models leads to the various Bayesian linear time series models (see Section 4.3.1). In general, random forests are algorithms that are based on simulation frameworks. Boosting algorithms are based on the optimization of appropriate loss functions, while neural networks seem to be more universal in the sense that implementations are expanded in all the categories. Obviously, exceptions to this general pattern exist, as it is shown in Figure 5.

| | Simulation – based methods | Point prediction methods | Distributional regression |
|---|---|---|---|
| Linear models | X (4.1) | X (4.2) | X (5.2) |
| Gaussian process regression | X (5.1) | | |
| Random forests | X (5.1) | | X (5.1) |
| Boosting algorithms | | X (5.4) | X (5.4) |
| Neural networks | X(6.2) | X (6.3) | X (6.4) |
| Other models | | X (5.5) | X (5.5) |

Figure 5. Primary algorithms for probabilistic prediction and a classification of them. The numbering in parentheses refers to the section, in which the description of the method can be found.

Questions about which scoring function or scoring rule to use to assess a point or probabilistic prediction often arise. A consistent scoring function should be used to assess predictions of a specified elicitable functional. However, there may exist a family of consistent scoring functions for the same functional. In this case, it is desirable to disclose the scoring function to the modeller, as consistent scoring functions for the same elicitable functional may provide different assessments. Under three assumptions, the choice of the consistent scoring function will not affect the assessment (Patton 2020):

(a) The information sets of the modellers are nested and do not lead to optimal predictions that are identical.

(b) The predictions are based on models that are perfectly estimated.

(c) The models are correctly specified for the functional of interest.

Unfortunately, these assumptions are rarely met in practice.

Even if the expected score is minimized, the choice of the scoring rule will affect the assessment results in finite samples, unless the three assumptions mentioned above are met (Patton 2020). In summary, the choice of the scoring function or scoring rule is an important consideration when assessing predictions. The scoring function should be disclosed to the modeller, and the assumptions underlying the scoring function should be carefully considered.

## 10.2 Differences between point prediction, distributional regression and Bayesian methods

Knowing the merits of each modelling approach is essential for deciding whether to use models based on consistent scoring functions or models based on distributional regression, when probabilistic predictions are required. Our considerations follow Waldmann (2018), who supports the use of quantile regression in specific situations. They also follow Rigby et al. (2013), who compare quantile regression with GAMLSS.

When one is interested in a specific elicitable functional, then he/she can apply a machine learning algorithm trained with a strictly consistent scoring function for this functional. That is e.g. the case in probabilistic forecasting competitions where a usually scaled quantile scoring function is disclosed to the competitors. In such competitions, machine learning quantile regression algorithms (e.g. deep learning (Section 6.3), boosting (Section 5.4), quantile regression forests (Section 5.3), support vector regression (Section 5.5.4)) are usually among the most successful ones. Predicting the functional is also possible using distributional regression. In this case, the predictive distribution is estimated and then it is straightforward to compute the predictive functional of interest. One can also use Bayesian statistical modelling.

A possible problem arising when using either distributional regression or Bayesian statistical modelling, is the misspecification of probability distributions. Indeed, in most practical situations, the probability distribution of the dependent variable is not known, while setting a probability distribution imposes strong assumptions. In some fields, such as the environmental sciences, some large-scale studies give guidance on the type of the most suitable distribution to model specific variables, although it is still possible that in many sub-cases of the problem at hand, this distribution may not be useful. One could overcome this limitation by using distributions that are more flexible in the sense of modelling more aspects of the dependent variable using many parameters. On the other hand, point prediction algorithms do not encounter this problem, since they are non-parametric and, thus, the additional assumption of probability distribution specification does not limit their flexibility.

On the other hand, the flexibility of point prediction algorithms may be prohibitive in cases that the available data are limited. In those cases, the algorithm may fail to converge rapidly. Therefore, some additional prior knowledge on the probability distribution may be exploited at the cost of reduced flexibility of the algorithm.

Another problem arising mostly in cases of simulation (e.g. Bayesian statistical) models, is the computational and storage cost. In such cases, one should simulate the full distribution that may be computationally prohibitive while also keeping the full sample for future use. That may be impractical if a single functional is of interest, in which case point prediction methods should be preferred. When the size of the data is small, then Bayesian modelling may become more appealing with respect to this fact. Progress in approximate Bayesian computation may also improve the practicality of Bayesian statistical models, while in some cases, part of the calculations is based on explicit formulas that can effectively reduce the need for additional computations. Unfortunately, in the latter case some additional assumptions may also be needed. A relevant example is the use of reference priors in Bayesian statistical modelling.

The size of the data is also important when one estimates high or extreme quantiles. In small data size cases, point prediction methods are not efficient for predicting such quantities, as it has been proven by large-sample theory. Distributional regression approaches might be preferable, although point-prediction algorithms can be adapted to deal with such cases. On the other hand, big data may also allow a suitable specification of the predictive distribution.

When the full predictive distribution of the dependent variable is of interest, Bayesian modelling and distributional regression are natural choices. However, it also possible to estimate functionals, such as quantiles at multiple levels, thereby estimating a substitute of the predictive distribution. That approach eliminates some of the advantages of point prediction methods regarding the speed of calculation, while quantile crossing becomes a possibility. Some advances remedy those undesirable properties, including for instance the simultaneous estimation of multiple quantiles using deep learning models.

In time series forecasting, the application of Bayesian modelling techniques is even harder, due to the larger number of parameters that are needed to be modelled and also due to the imposition of dependence structure. However, again some time series may be short. In this latter case, Bayesian modelling might be benefitting.

Scoring functions for all tasks have not been discovered yet, while the dependent variable may be characterized by some peculiarities, with intermittency consisting a characteristic example of the latter. In such cases, distributional regression might be more suitable. Indeed, for the example of intermittency, modelling can be effective when using appropriate probability distributions (Ziel 2021).

Tradition in point prediction dictates that different models are assessed in a test (out-of-sample) set and the modeller selects the model with the better performance. A scoring function that is consistent for the functional of interest is used to rank the models. That cross-validation-based tradition is rarely met in Bayesian statistical modelling and to a lesser extent in distributional regression. That is reasonable, since Bayesian statistical modelling mostly focuses on parameters inference. Comparison of different models is based on prediction coverages and widths of predictions intervals, although those scoring rules are not proper.

## 10.3 Future outlook

Several challenges for advancing the field of probabilistic prediction are related to tasks that existing algorithms are not able to complete. These tasks include but are not limited to:

**Algorithms tailored to client's requirements**: Existing algorithms are mostly trained with prespecified scoring functions or scoring rules. Those scoring rules may not meet users' needs. For instance, estimation of quantiles may not be adequate for quantifying the impact of exceedance, while the magnitude of the exceedance may also be of interest. New scoring functions may be properized to be made consistent for functionals. Optimization of algorithms using proper scoring rules beyond local ones is also a strategy for obtaining improved probabilistic predictions.

**Temporal and spatial predictions**: Although models for time series forecasting and spatial prediction have been developed, several topics have place for improvements. Those include, for example, the prediction of multiple points simultaneously. Related scoring rules for adapting algorithms exist (e.g. the energy score); however, these are rarely met in practice. The task is more challenging compared to the case of predicting the mean of multiple variables simultaneously. For instance, a problem that has been rarely addressed is that of predicting multiple quantiles, possibly at different levels, simultaneously.

**Prediction of extreme quantiles**: Predicting extreme quantiles of probability distributions is another challenge and hot topic of current research. The scoring functions for the task are not satisfying so far. Distributional regression may be an option, but is accompanied with several disadvantages (which were already mentioned). The current state of research seems to be heading to a direction that intermediate functionals are predicted and then they are extrapolated using extreme value theory. Other scoring functions that exploit information beyond the frequency over specified levels seem to also be a trending approach.

**Observational errors**: Modelling of observational errors seems to witness slow progress. Perhaps this is due to the lack of suitable large datasets. Progress seems to be heading towards the direction of developing new scoring functions, as well as towards obtaining progress in Bayesian statistical modelling.

**Large-scale comparisons**: While large scale comparisons are regularly met for the case of machine learning algorithms for point predictions and forecasts (with these predictions and forecasts being related to means or medians of the dependent variable), that does not seem to be the case for probabilistic predictions. Such comparisons may reveal properties of algorithms based on empirical evidence. When the task is forecasting, time series are of varying magnitudes; therefore, skill scores might be more suitable for comparing competing algorithms. However, such scores may not be proper. Development of statistical tests of the significance of the difference in the performance of different methods still witnesses some small progress. Combining multiple algorithms might also be a way for improving predictive performance and much is left to be desired towards this direction in the respective field.

**Other topics**: Other challenges are related to the proper scoring of prediction intervals. Historically, prediction intervals seem to be among the first functionals of interest that were predicted in time series forecasting. The usual strategy for comparing algorithms was based on a metric of calibration (specifically, on coverage probabilities) and on a metric of sharpness (specifically, on widths of prediction intervals). Although they can be informative, such comparisons are not consistent to the task of predicting the functional. Related scoring rules should be used; however, these scoring rules mostly refer to fixed quantile levels. Other approaches for scoring predictions of intervals with varying quantile levels should be developed. Another challenge is related to the prediction of dependent variables with peculiar properties, such as intermittency, multiple modes, heavy tails and more. New scoring rules may help for calibrating machine algorithms to achieve the specific task.

## 10.4 Final remarks

Although, the various applications of probabilistic predictions are increasing, the vast majority of the machine learning applications refer to point predictions of the mean functional of the distribution of the dependent variable. Due to reasons of tradition, as well as for making the communication of application products easier for their users, the current situation (which is in favour of point predictions) is not expected to change. However, we anticipate that the applications of probabilistic predictions using machine learning algorithms will increase their share, both in the academia and in the industry. We also anticipate that the procedure of users' decisions will adapt to the increased information provided by probabilistic predictions. Further progress in developing new algorithms is also expected to meet the increasing needs.

**Conflicts of interest:** The authors declare no conflict of interest.

**Author contributions:** HT and GP contributed equally to this work.

**Acknowledgements:** The authors are deeply grateful to the Editor for his careful handling of the review process and to the Reviewers for their insightful and constructive comments, which have resulted in a significantly improved article. The Reviewers' suggestions have helped us to expand the scope of the article, clarify the arguments, and improve the overall quality of the writing.